\documentclass[10pt,journal,compsoc]{IEEEtran}
\usepackage{graphicx}
\usepackage{subcaption}
\usepackage{enumitem}
\usepackage{hyperref}
\hypersetup{colorlinks=true,linkcolor=blue,anchorcolor=blue,citecolor=blue}
\usepackage{multirow}
\usepackage{setspace}
\usepackage{multicol}
\usepackage{stfloats}
\newcommand{\mathbfz}{\mathbf{z}}

\usepackage[table]{xcolor}
\definecolor{lightgray}{RGB}{220, 220, 220}
\usepackage{xspace}
\usepackage{bbding}
\usepackage{pifont}
\usepackage{ragged2e}
 \usepackage{makecell}
\newcommand{\etal}{\emph{et al.}\xspace}

\newcommand{\myblue}[1]{{\color{black}#1}}
\newcommand{\mygreen}[1]{{\color[rgb]{0,0.5,0}\textbf{#1}}}

\newcommand{\taskAA}{\textcolor{red}{\ding{170}}} %
\newcommand{\taskBB}{\textcolor{red}{\ding{169}}} %
\newcommand{\taskCC}{\textcolor{black}{\ding{168}}} %
\newcommand{\taskDD}{\textcolor{black}{\ding{171}}} %
\newcommand{\taskEE}{\textcolor{blue}{\ding{72}}} %

\usepackage{algorithm}
\usepackage{algorithmic}
\usepackage{amsmath}

\usepackage{amssymb}
\usepackage{booktabs}
\usepackage{multirow}

\begin{document}

\title{\huge Uni-Hand: Universal Hand Motion Forecasting\\ 
in Egocentric Views}%

\author{Junyi~Ma$^1$, Wentao~Bao$^2$, Jingyi~Xu$^3$, Guanzhong~Sun$^4$, Yu~Zheng$^1$, Erhang~Zhang$^1$, \\ Xieyuanli~Chen$^{5}$, Hesheng~Wang$^{1*}$
\thanks{This work was supported by National Key R\&D Program of China (Grant No.2024YFB4708900), and Shanghai
Municipal Science and Technology Major Project (Grant No.2025SHZDZX025G01). It was also supported in part by the Natural Science Foundation of China under Grants 62225309, U24A20278, 62361166632, and U21A20480. $^{1}$Junyi~Ma, Yu~Zheng, Erhang~Zhang, and Hesheng~Wang are with IRMV Lab, Shanghai Jiao Tong University. $^{2}$Wentao~Bao is with Meta Reality Labs. $^{3}$Jingyi~Xu is with the Department of Electronic Engineering, Shanghai Jiao Tong University. $^{4}$Guanzhong~Sun is with China University of Mining and Technology. $^{5}$Xieyuanli~Chen is with the College of Intelligence Science and Technology, National University of Defense Technology.}
\thanks{$^{*}$Corresponding author emails: wanghesheng@sjtu.edu.cn}

}

\IEEEtitleabstractindextext{
\begin{abstract}
Forecasting how human hands move in egocentric views is critical for applications like augmented reality, human-robot policy transfer, and service/assistive technologies. Recently, several hand trajectory prediction (HTP) methods have been developed to generate future possible hand waypoints, which still suffer from insufficient prediction targets, inherent modality gaps, entangled hand-head motion, and limited validation in downstream tasks.
To address these limitations, we present Uni-Hand, a universal hand motion forecasting framework considering multi-modal input, multi-dimensional and multi-target prediction patterns, and multi-task affordances for downstream applications. We harmonize multiple modalities by vision-language fusion, global context incorporation, and task-aware text embedding injection, to forecast hand waypoints in both 2D and 3D spaces.
A novel dual-branch diffusion is proposed to concurrently predict human head and hand movements, capturing their motion synergy in egocentric vision. By introducing target indicators, the prediction model can forecast the specific joint waypoints of the wrist or the fingers, besides the widely studied hand center points. In addition, we enable Uni-Hand to additionally predict hand-object interaction states (contact/separation) to facilitate downstream tasks better. \myblue{To incorporate comprehensive downstream task evaluations in the literature, we build novel benchmarks to assess the real-world applicability of hand motion forecasting algorithms.}
The experimental results on multiple publicly available datasets and our newly proposed benchmarks demonstrate that Uni-Hand achieves the state-of-the-art performance in multi-dimensional and multi-target hand motion forecasting. Extensive validation in multiple downstream tasks also presents its impressive human-robot policy transfer to enable robotic manipulation, and effective feature enhancement for action anticipation/recognition. We will release our code, pretrained models of Uni-Hand, and novel benchmarks at the project page: \url{https://irmvlab.github.io/unihand.github.io}.
\end{abstract}


}

\maketitle
\IEEEdisplaynontitleabstractindextext
\IEEEpeerreviewmaketitle

\section{Introduction}
\label{sec:intro}

Predicting human behaviors over future time horizons is critical for applications such as augmented reality and human-robot interaction. As an important interface for human-environment interaction, the human hand offers valuable motion knowledge that facilitates various downstream tasks in embodied intelligence~\cite{qiu2025humanoid,kareer2024egomimic,gong2020rcar,bandini2020analysis}. However, despite significant progress in other human behavior anticipation tasks like
predicting gaze locations~\cite{zhang2017deep,lai2024eye,li2018rcar,li2018eye}, and interaction regions~\cite{roy2024interaction,nagarajan2020ego,mur2024aff}, forecasting fine-grained human hand motion still remains challenging. 
To align better with hand movement patterns and potential intentions from a human perspective,
some egocentric-vision-only hand motion forecasting (HMF) approaches~\cite{liu2020forecasting,liu2022joint,bao2023uncertainty,ma2024diff} have been developed in recent years. They receive past human egocentric (first-person) observations and directly predict future hand trajectories in the specified reference frame. However, these methods exhibit significant limitations that hinder their further performance improvement and effectiveness in possible downstream applications:
\begin{figure}[t]
  \centering
  \includegraphics[width=0.87\linewidth]{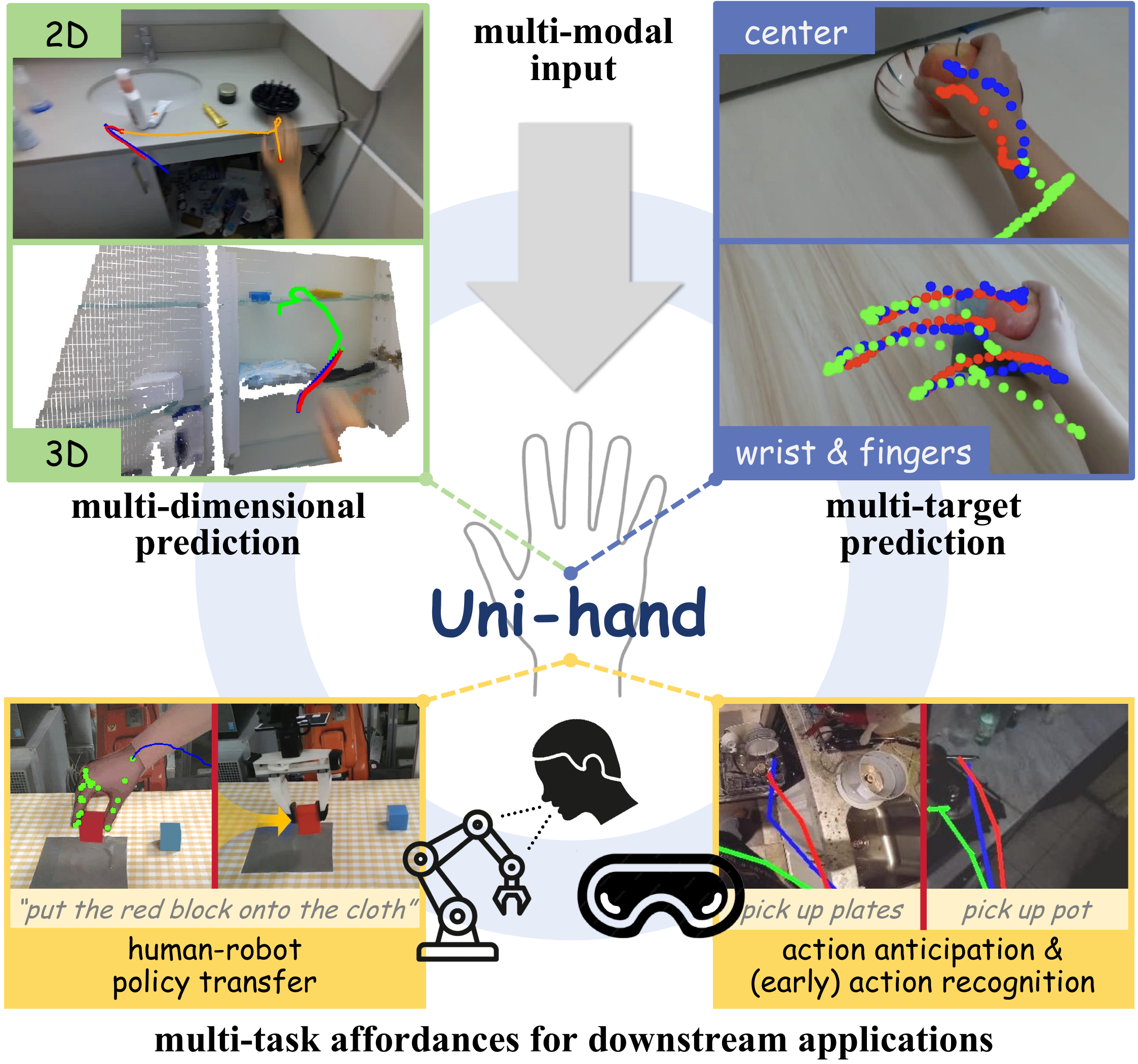}
  \vspace{-0.3cm}
  \caption{Uni-Hand is a universal hand motion forecasting framework which facilitates multi-dimensional and multi-target predictions with multi-modal input. It also enables multi-task affordances for downstream applications.}
  \label{fig:motivation}
  \vspace{-0.5cm}
\end{figure}
\begin{itemize}[leftmargin=1em]
\item \textbf{Insufficient prediction targets:} The existing hand motion forecasting paradigms exclusively concentrate on the generation of future hand centers (of the potential bounding boxes) as target motion waypoints~\cite{bao2023uncertainty,ma2024diff,ma2024madiff}.
Despite its necessity for providing hand movement trends, these methods basically overlook how different joints of the fingers and wrist move in 2D and 3D spaces. Besides, they do not involve locating the precise timings when the hand will make contact with and separate from target objects. Therefore, there is a lack of sufficient motion knowledge in future time horizons for practical deployments, such as human-robot manipulation policy transfer.

\item \textbf{Modality gaps:} \myblue{Existing HMF paradigms predominantly operate on 2D egocentric video inputs, without explicitly incorporating metric-scale 3D spatial observations.
While some prior works have shown that temporal modeling, pose estimation, and learned priors can partially infer potential 3D dynamics/structural information from 2D inputs~\cite{ma2024madiff,ma2025eer,hatano2024emag,zhou2025megohand}, relying only on 2D visual cues without access to raw 3D perception may introduce geometric ambiguity in fine-grained 3D motion forecasting and physically grounded interactions.} In addition, the absence of textual modality interfaces prevents explicitly incorporating known human intentions or instructions. It also exacerbates the discrepancy between the HMF model optimization and possible deployments in language-conditioned downstream tasks.

\myblue{\item \textbf{Lack of prediction-level head motion conditioning:}  
Human hand motion is closely linked to whole-body kinematics, where the head movement is its most straightforward concurrent motion to provide visual observations. Understanding how the head moves benefits hand motion forecasting. While prior works~\cite{ma2024diff,ma2024madiff,hatano2024emag} encode \textit{past} headset camera egomotion into hand state transition to couple head and hand features, they do not explicitly forecast \textit{future} head motion to explicitly condition hand motion prediction. This impairs the prediction models' ability to comprehend future hand motion patterns in the egocentric view of incoming hand-head cooperation.}

\item \textbf{Limited validation in downstream tasks:}
The existing HMF frameworks aim to reduce the errors between predicted waypoints and ground-truth human demonstrations. While this objective is well-justified, their capability to support downstream tasks should be further analyzed in this literature. \myblue{Transferring human actions to robotic manipulation has captured substantial attention nowadays owing to the high accessibility of human demonstrations~\cite{qiu2025humanoid,li2025maniptrans}. In line with this trend, evaluating HMF models via robotic manipulation is particularly natural, as the predicted human hand trajectories from videos can be directly mapped to robot end-effector action planning, providing an interpretable HMF assessment for physical plausibility.
Therefore, exploring the transferability of HMF models to robotic manipulation tasks warrants significant research attention. 
In addition, predicted hand motion contains rich human behavior semantics, which are strongly correlated with action categories. In contrast to designing task-specific ad-hoc decoders~\cite{liu2020forecasting,liu2022joint}, how predicted hand motion features directly enhance the existing framework of downstream action category anticipation and recognition needs to be extensively explored.}

\end{itemize}

To address the above limitations, we propose a universal hand motion forecasting framework, dubbed \textit{Uni-Hand}. Uni-Hand absorbs multi-modal inputs to achieve multi-dimension/-target hand motion prediction, and effectively provides multi-task affordances for downstream applications. \textbf{To extend the prediction targets}, Uni-Hand introduces target indicators in denoising diffusion to forecast hand centers or specified joints in both 2D and 3D space. It also integrates an additional interaction state decoder to locate the timings of hand-object contact and separation events. We therefore present a more flexible HMF solution compared to the existing paradigms~\cite{liu2022joint,ma2024diff,hatano2024emag} that only attend to hand center prediction. \textbf{To narrow the modality gaps}, we incorporate multiple modalities as the inputs including 2D RGB images, 3D point clouds, past hand waypoints, and text prompts. This not only enhances the awareness of 3D structure information in our framework, but also provides a textual modality interface for integrating possible language instructions in real-world robotic applications. \myblue{\textbf{To address the lack of prediction-level head motion conditioning}, we design a dual-branch diffusion to concurrently forecast head and hand movements, explicitly capturing their synergy for more accurate HMF.} Ultimately, \textbf{for extensive validation in downstream tasks}, we deploy our framework on multiple real-world robotic tasks, validating its capability to facilitate atomic skills, language-conditioned implementations, and long-horizon tasks. We also demonstrate its affordable motion features to improve the off-the-shelf framework for conventional action anticipation, early action recognition, and action recognition tasks.

In summary, the \textbf{main contributions} of this paper are fourfold:
\begin{itemize}[leftmargin=1em]
\item \textbf{Universal HMF framework:} We present Uni-Hand, a universal hand motion forecasting framework, featuring a versatile solution for multi-dimension/-target HMF and multi-task affordances for downstream applications. 

\item \textbf{Dual-branch diffusion:} We propose a novel dual-branch diffusion, which employs multi-modal information as input to concurrently predict future headset camera egomotion and hand movements in egocentric views. 

\item \textbf{Hybrid Mamba-Transformer denoising:} A novel hybrid Mamba-Transformer module is designed for denoising hand motion latents. It combines the strong temporal modeling ability of Mamba and the global context-awareness of Transformer to harmonize multi-modal features. We also inject text embeddings into the hybrid module to facilitate language-conditioned tasks.

\item \textbf{HMF benchmark:} To the best of our knowledge, this is the first HMF research considering the assessment of hand forecasting algorithms' effectiveness in enabling comprehensive downstream tasks, including robotic manipulation, action anticipation, early action recognition, and action recognition. The extensive experiments on multiple publicly available datasets and our newly proposed benchmarks demonstrate that Uni-Hand achieves SOTA performance on 2D/3D hand motion forecasting, and can effectively adapt to the downstream tasks.

\end{itemize}

This work is an extension of the preliminary version~\cite{ma2025mmtwin},
in which we only explored how to harmonize multiple input modalities for hand center prediction and achieve decoupling hand-head motion forecasting. In this study, we substantially expand in the
following ways:
\begin{itemize}[leftmargin=1em]
\item We extend the original 3D prediction paradigm to a universal framework for multi-dimension/-target HMF and multi-task affordances for downstream applications.

\item We incorporate additional prediction targets besides hand centers, encompassing multiple hand joints and interaction states (contact/separation).

\item A new textual modality interface (text embedding injection) is designed to enhance our model's awareness of specific task instructions. 

\item A comprehensive extension of experiments on public datasets and our new benchmarks presents the model's foundation capability to facilitate downstream tasks including real-world robotic manipulation, action anticipation, early action recognition, and action recognition.
\end{itemize}

This paper is organized as follows. Sec.~\ref{sec:related_work} reviews related works in literature. Sec.~\ref{sec:unihand} details Uni-Hand's architecture and training/inference schemes. Sec.~\ref{sec:exp} showcases the extensive experimental results quantitatively and qualitatively. \myblue{Sec.~\ref{sec:discussion} highlights our algorithmic novelty and provides our insights.} Ultimately, Sec.~\ref{sec:conclusion} concludes the paper.

\vspace{-0.1cm}

\section{Related Work}
\label{sec:related_work}

\subsection{Hand Trajectory Prediction in Egocentric Vision}
\label{sec:htp_related_work}

Hand trajectory prediction (HTP) is a key component in understanding future hand motion patterns.
Given sequential egocentric observations, HTP aims to forecast future possible hand waypoints in the 2D image plane or 3D space. Here, the hand waypoints denote the bounding box centers of detected hands. 
Motivated by the need for scaling capabilities, some 2D HTP approaches have been pioneeringly developed based on a large volume of human activity videos. For example, OCT~\cite{liu2022joint} uses a Transformer~\cite{vaswani2017attention} based encoder-decoder architecture to jointly predict hand waypoints and object hotspots. In contrast, Diff-IP2D~\cite{ma2024diff} replaces the Transformer blocks in OCT with egomotion-aware diffusion models, achieving better 2D HTP performance. Similar to Diff-IP2D, EMAG~\cite{hatano2024emag} also innovatively encodes homography matrices of camera egomotion into 2D trajectory prediction with the cross-attention mechanism. These two works make initial efforts to integrate headset camera egomotion awareness into the 2D HTP pipeline. To further enhance the interpretability regarding fused camera egomotion, MADiff~\cite{ma2024madiff} incorporates Mamba~\cite{gu2023mamba} to model sequential hand state transition, based on a new motion-driven selective scan. Its powerful egomotion-aware Mamba block is out-of-the-box with respect to 2D predictive tasks in first-person views. More recently, HandsOnVLM~\cite{bao2024handsonvlm} introduces a VLM scheme to transform 2D HTP into a VQA-style task. 
To further validate the applicability in real-world scenes, Bahl~\etal~\cite{bahl2023affordances} build an affordance model and deploy it for robot data collection and reward-free exploration.

Although 2D HTP is more scalable due to the availability of internet-scale human videos~\cite{ma2024madiff}, 3D HTP yields more human-intent-aligned results. For instance, USST~\cite{bao2023uncertainty} receives image prompts to directly generate future 3D hand waypoints by autoregressive Transformers. EgoH4~\cite{hatano2025invisible} extends to body pose estimation, predicting locations of invisible hands. In contrast to EgoH4 focusing on whole body movements, our recent MMTwin~\cite{ma2025mmtwin} attends to the most critical body part, i.e., the head and hand, in hand-object interaction, forecasting their future motion concurrently. Compared to egomotion-aware 2D HTP~\cite{ma2024diff,hatano2024emag,ma2024madiff} that only encodes past head motion, MMTwin builds an additional diffusion branch to predict future head motion latents. 

Despite the significant progress in the above-mentioned 2D and 3D hand trajectory prediction, these methods basically only determine future centers of potential hand bounding boxes. However, the hand center constitutes an excessively coarse-grained prediction target, incapable of representing hand pose variations. This hinders fine-grained understanding of human behavior in hand-object interaction. Besides, these methods are agnostic to the potential timings of hand-object contact and separation. Therefore, they cannot be easily transferred to downstream tasks such as robot manipulation, since trajectories can only determine ``where to move'' rather than "when to grasp". In contrast, Uni-Hand proposed in this work can forecast future 2D and 3D positions of multiple hand joints besides center points, leading to awareness of hand pose changes in egocentric views. We introduce target indicators to let the denoising diffusion model distinguish different prediction targets. Moreover, we develop an additional decoder to forecast future interaction states, which can determine the hand-object contact/separation periods. Thus, Uni-Hand extends the conventional HTP literature to more informative hand motion forecasting by enriching prediction targets, and can practically facilitate multiple downstream applications.

\vspace{-0.2cm}

\subsection{Diffusion Models in Hand Motion Analysis}
\label{sec:dm_related_work}

In recent years, hand motion analysis has attracted much attention due to the informational richness of hand-centric data for downstream tasks. While object-agnostic hand mesh recovery~\cite{pavlakos2024reconstructing,dong2024hamba,Li_2024_CVPR,tian2023recovering} and object-aware 3D HOI reconstruction~\cite{on2025bigs,ye2022s,liu2024easyhoi,zhu2023get} have witnessed significant advances, synthesizing reasonable hand poses around target objects and analyzing hand dynamics along the time axis are more challenging. With the advent of diffusion models~\cite{ho2020denoising,dhariwal2021diffusion}, some works such as HOIDiffusion~\cite{zhang2024hoidiffusion}, Gears~\cite{zhou2024gears}, and DiffH2O~\cite{christen2024diffh2o} seamlessly integrate the powerful generation ability of diffusion models to generate hand meshes from scratch. However, they basically underestimate how the hand approaches the target object. More recently, PEAR~\cite{zhang2024pear} considers analyzing the hand motion trend during the interaction process. In contrast to PEAR, Tang~\etal~\cite{tang2024prompting} develop a prompt-based future-driven diffusion model for predicting future 3D coordinates of multiple hand joints. Despite being more fine-grained, this method does not attend to wide-range wrist movements. Conversely, Diff-IP2D~\cite{ma2024diff} and MADiff~\cite{ma2024madiff} both utilize DiffuSeq~\cite{gong2022diffuseq}-style models to forecast future wide-range hand movements, but lose detailed joint motion as mentioned before. 

This work also adapts the advanced generation ability of diffusion models, to achieve multi-dimensional and multi-target future hand motion forecasting. Unlike the existing diffusion-based schemes, we propose a novel hybrid Mamba-Transformer architecture as the denoising model to harmonize multi-modal input. Besides, we introduce camera egomotion analysis within the dual-branch diffusion to enable more comprehensive hand motion understanding.

\begin{figure*}[t]
  \centering
  \includegraphics[width=0.95\linewidth]{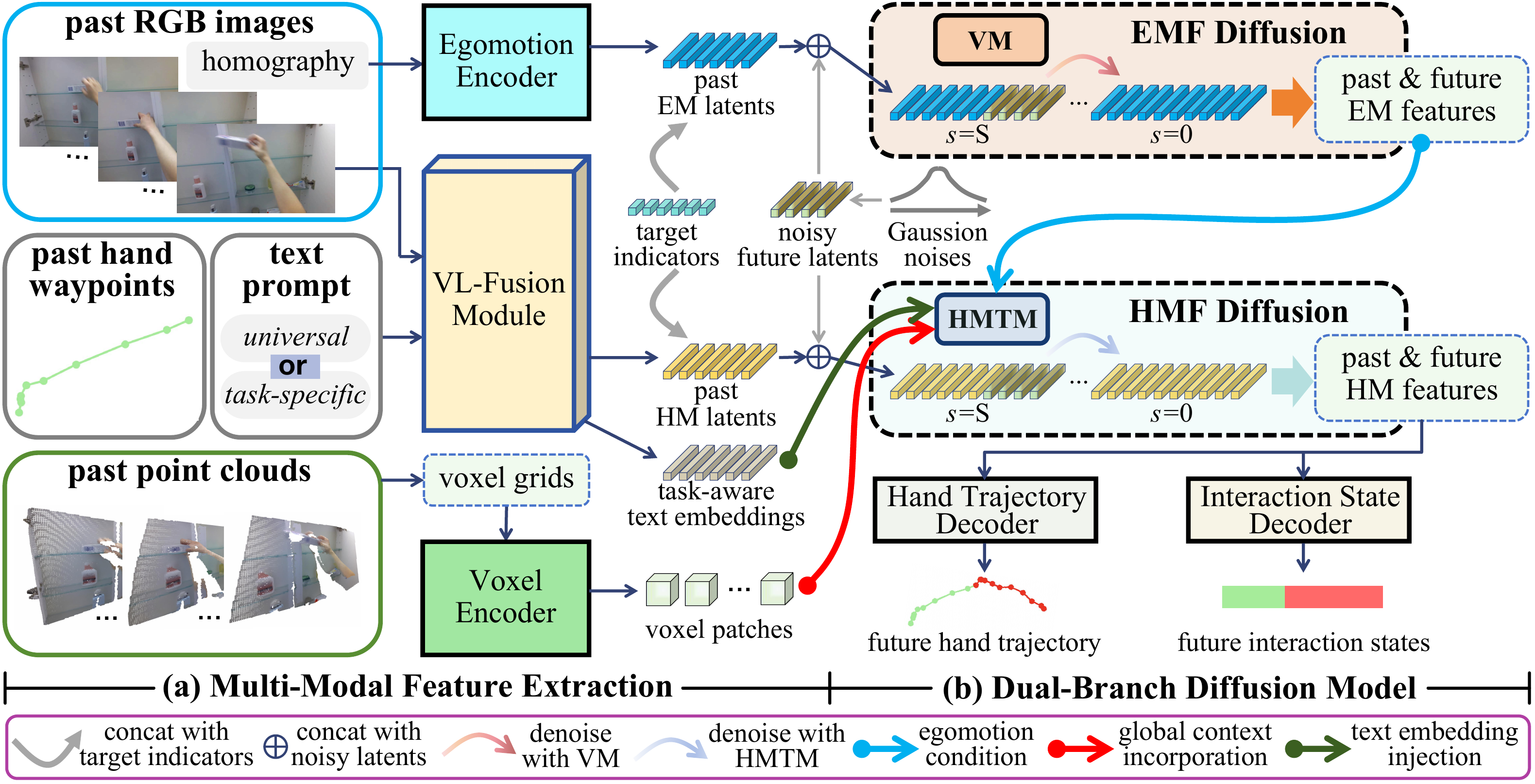}
  \caption{System overview of Uni-Hand. Uni-Hand (a) converts multi-modal input into latent feature spaces, and (b) decouples predictions of future egomotion latents (EM latents) and hand motion latents (HM latents) by a novel dual diffusion. The vanilla Mamba (VM) is used for denoising in the ego-motion-forecasting diffusion (EMF diffusion). We further design a new denoising model in hand-motion-forecasting diffusion (HMF diffusion) with a hybrid Mamba-Transformer module (HMTM). The predicted HM latents are ultimately decoded to future hand trajectories and interaction states.}
  \label{fig:system_overview}
  \vspace{-0.55cm}
\end{figure*}

\vspace{-0.2cm}

\subsection{Downstream Tasks of Hand Motion Forecasting}
\label{sec:hrp_related_work}

As hands are fundamental to physical interaction in human daily life, they have driven growing research interest in human-robot policy transfer for manipulation tasks. Traditional imitation learning~\cite{chi2023diffusion,ze20243d} techniques rely on costly teleoperation and lack cross-embodiment generalization. In contrast, widely available human videos already contain rich demonstrations. They are robot-agnostic and offer transferable knowledge for manipulation. Therefore, directly capturing human hand motion data from egocentric or exocentric videos to guide robot policy generation is a promising path towards embodied intelligence. Compared to producing pretrained visual representations from large-scale human videos~\cite{nair2022r3m,majumdar2023we,xiao2022masked,pramanick2023egovlpv2}, extracting explicit affordances related to human hands enables more straightforward human-robot policy transfer. For example, OKAMI~\cite{li2024okami} retargets the body motions and hand poses separately from a single video imitation to humanoid robot actions. Following OKAMI using only one video demonstration, YOTO~\cite{zhou2025you} proposes a proliferating strategy for one-shot teaching, enabling more demonstrations for the bimanual diffusion policy. Instead of relying entirely on human video data, Motion Tracks~\cite{ren2025motion} combines robot and human video demonstrations to train a motion prediction network. Compared to these imitation learning schemes, ManipTrans~\cite{li2025maniptrans} models the transfer process with reinforcement learning.
Wen~\etal~\cite{wen2023any} design a two-stage framework for any-point trajectory modeling, which guides policy learning with predicted future point trajectories. More recently, Qiu~\etal~\cite{qiu2025humanoid} unify the state-action space of humanoid embodiments and human hands for action prediction. 

Inspired by these human-robot transfer strategies, in this work, we expand the existing HMF assessment paradigm~\cite{liu2020forecasting,liu2022joint,ma2024diff} by evaluating Uni-Hand's capability to support robot manipulation tasks. That is, we directly use pretrained Uni-Hand to generate executable trajectories and gripper actions for a real robot. The success rates of multiple manipulation tasks highlight its potential for real-world downstream deployments. In addition, we also present whether Uni-Hand can benefit the other three critical downstream tasks in egocentric vision, i.e., action anticipation~\cite{mittal2024can,qi2024uncertainty,zhong2023anticipative,furnari2020rolling}, early action recognition~\cite{stergiou2023wisdom,weng2020early,furnari2020rolling}, and action recognition~\cite{perrett2021temporal,zheng2020dynamic,wang2024tamt,hatano2024multimodal,furnari2020rolling}. The improvement of anticipation and recognition accuracy by introducing hand motion features shows Uni-Hand's applicability in possible wearable assistive technologies.

\section{Proposed Method}
\label{sec:unihand}

\subsection{System Overview}
\label{sec:unihand_archtecture}

Here we first provide the overall inference pipeline of Uni-Hand in Fig.~\ref{fig:system_overview}. 
Given a sequence of past egocentric observations $\mathcal{O}=\{O_t\}_{t=-N_\text{p}+1}^{0}$, past waypoints of the hand center or a specific joint $\mathcal{H}_\text{p}=\{H_t\}_{t=-N_\text{p}+1}^{0} \,(H_t \in \mathbb{R}^{2} \,\text{or}\, \mathbb{R}^{3})$, and a text prompt $L$, our proposed Uni-Hand aims to predict future hand trajectories $\mathcal{H}_\text{f}=\{H_t\}_{t=1}^{N_\text{f}} (H_t \in \mathbb{R}^{2} \,\text{or}\, \mathbb{R}^{3})$, as well as interaction states $\mathcal{G}^\text{f}=\{G_t\}_{t=1}^{N_\text{f}} (G_t \in \{0,1\})$ if needed. $N_\text{p}$ and $N_\text{f}$ denote the number of frames in the past and future time horizons. Each observation $O_t$ consists of an egocentric 2D RGB image $I_t \in \mathbb{R}^{c\times h\times w}$ and 3D point clouds $D_t \in \mathbb{R}^{n\times 3}$. The temporal interaction state $G_t=0$ indicates that the hand and target object are physically separated, while $G_t=1$ represents that they are in contact. 
Following the previous works \cite{liu2022joint,bao2023uncertainty,ma2024diff}, we forecast the future hand waypoints on a fixed reference frame (e.g., the image plane and the camera coordinate system of the first frame $O_{-N_\text{p}+1}$ for 2D and 3D HMF tasks, respectively). As can be noted, our framework absorbs multi-modal information to enhance the prediction model's perceptual capability of environments and tasks, compared to the current baselines~\cite{liu2020forecasting,liu2022joint,ma2024diff,hatano2024emag} only using image inputs. Besides, UniHand can also predict waypoints for multiple joints (e.g., $\mathcal{H}_\text{f}^{j0}$, $\mathcal{H}_\text{f}^{j4}$, and $\mathcal{H}_\text{f}^{j8}$) in parallel since the data from different joints can be packed into a single batch.

Specifically, we first extract features/latents from the input RGB images, hand waypoints, and text prompt (Fig.~\ref{fig:system_overview}(a), Sec.~\ref{sec:mmfe}). 
The VL-fusion module generates hand motion latents (HM latents) by fusing vision-language features, waypoint features, task-aware text embeddings, and target indicators. The egomotion encoder computes sequential homography matrices of the input image sequence as headset camera egomotion, and then encodes them to egomotion latents (EM latents). To integrate 3D structure features, the voxel encoder converts the past point clouds to voxel patches. Subsequently, a dual-branch diffusion model (Fig.~\ref{fig:system_overview}(b), Sec.~\ref{sec:dbdm}), encompassing hand-motion-forecasting diffusion (HMF diffusion) and ego-motion-forecasting diffusion (EMF diffusion), is tailored for concurrently predicting future HM latents and EM latents conditioned on their past counterparts. The EMF diffusion uses the vanilla Mamba to efficiently denoise the noisy future latents. In contrast, the HMF diffusion adopts our hybrid Mamba-Transformer module as the denoising model, which consists of stacked egomotion-aware Mamba blocks and structure-aware/task-aware Transformer.
It harmonizes hand motion modeling for egocentric vision, text embedding injection for task-aware optimization, and global context incorporation for 3D structure awareness. Ultimately, the denoised future HM latents are decoded to future hand trajectories and interaction states. We train the components of our Uni-Hand in an end-to-end manner with multiple loss functions (Sec.~\ref{sec:training_and_inf}).

\subsection{Multi-Modal Feature Extraction}
\label{sec:mmfe}

\subsubsection{Egomotion Encoder}
\myblue{Since our framework considers modeling head motion in hand motion forecasting, we first calculate sequential homography $\mathcal{M}_\text{p}=\{M_t\}_{t=-N_\text{p}+1}^{0} \,(M_t \in \mathbb{R}^{3\times 3})$ from ${I_t}$ to represent past headset camera egomotion following~\cite{ma2024diff,hatano2024emag}. $M_t$ denotes the homography matrix between $t\,\text{th}$ frame and the first frame $(t=-N_\text{p}+1)$ estimated by SIFT descriptors~\cite{lowe2004distinctive} with RANSAC \cite{fischler1981random}. Subsequently, the egomotion encoder, composed of MLPs, encodes egomotion homography $\mathcal{M}$ to past egomotion features $F^\text{em}_\text{p}\in \mathbb{R}^{N_\text{p}\times f}$ as latents for the following EMF diffusion. Note that the ground-truth (GT) future EM latents $F^\text{em}_\text{f}\in \mathbb{R}^{N_\text{f}\times f}$ are obtained by encoding the GT homography matrices $\mathcal{M}_\text{f}=\{M_t\}_{t=1}^{N_\text{f}}$ computed from future image frames using the same egomotion encoder as for the past frames. They only exist in the training process for supervision and will be replaced by sampled noises $F^\text{em}_\text{noise}\in \mathbb{R}^{N_\text{f}\times f}$ to enable denoising-based inference in EMF diffusion as shown in Fig.~\ref{fig:system_overview}(b).}

\begin{figure}[t]
  \centering
  \includegraphics[width=1\linewidth]{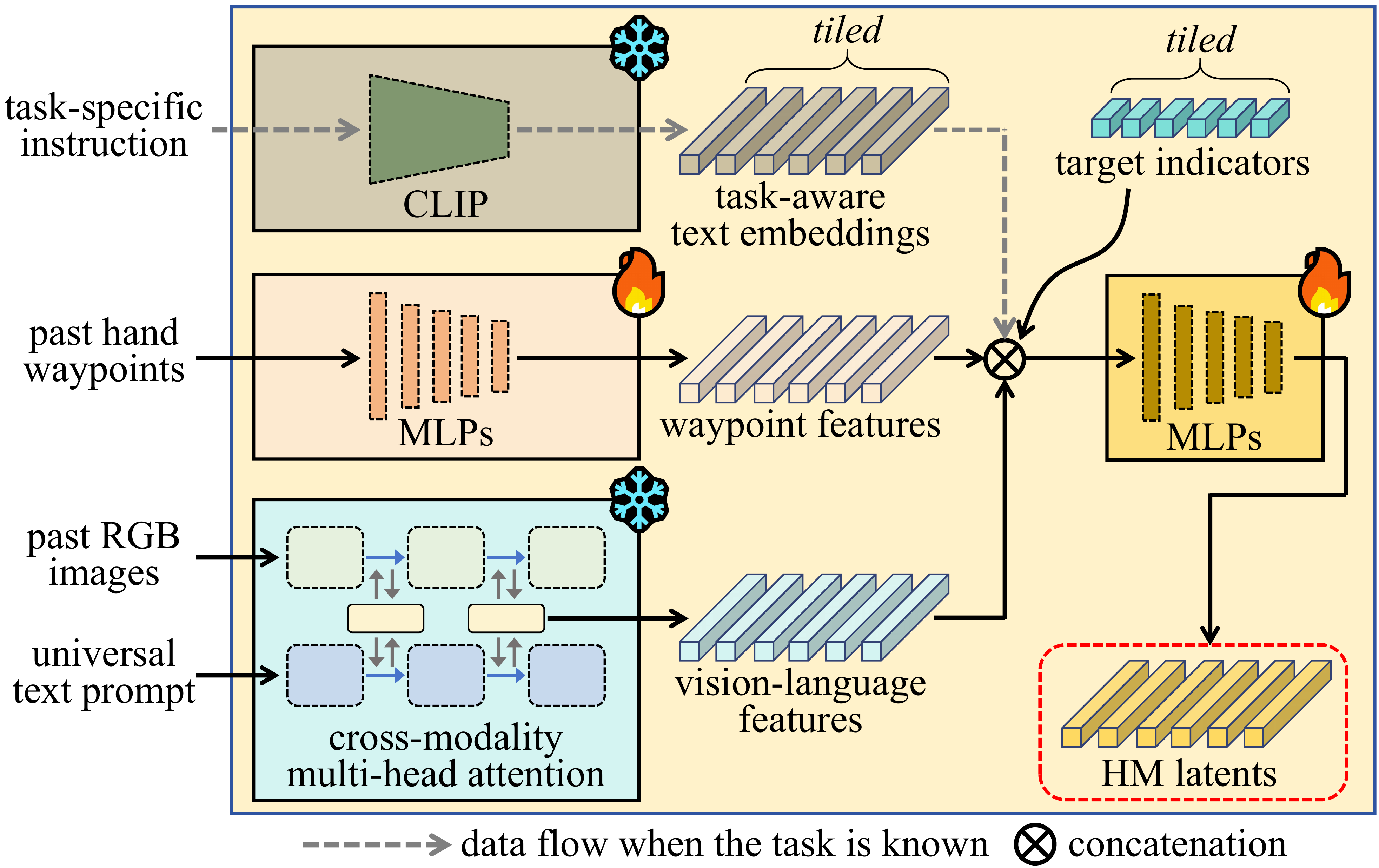}
  \caption{Architecture of the VL-fusion module. It generates HM latents for the following HMF diffusion by fusing vision-language features, waypoint features, and task-aware text embeddings.}
  \label{fig:fusion_module}
  \vspace{-0.5cm}
\end{figure}

\subsubsection{VL-Fusion Module}
\label{sec:vfm}
For hand motion forecasting, we develop the VL-fusion module to generate HM latents by fusing multiple modalities, including RGB images, past waypoints, and text prompts. As shown in Fig.~\ref{fig:fusion_module}, here we support two types of text prompting for different HMF applications, the universal prompt and the task-specific instruction. When the target hand motion is task-agnostic, we advocate using the universal prompt ``\textit{hand}'' to generate vision-language fusion features $X^\text{vl}\in \mathbb{R}^{(N_\text{p}+\ell)\times x}$ by the pretrained GLIP~\cite{Li_2022_CVPR} following~\cite{ma2024madiff}. $x$ is the feature channel dimension. $\ell$ equals $N_\text{f}$ during training, and is set to $0$ during inference since future HM latents will be replaced by sampled noises in the inference process of our HMF diffusion. The universal text prompt helps to semi-implicitly capture hand poses and
hand-scenario relationships within each image through GLIP's deepest cross-modality multi-head attention.
Conversely, once the task being performed by the hand is known, an additional task-specific instruction (e.g., ``\textit{push the block onto the cloth}'') is absorbed by the VL-fusion module to generate the text embedding $X^\text{task}\in \mathbb{R}^{x}$ with CLIP~\cite{radford2021learning} for task-awareness in hand motion forecasting. 

Moreover, we use MLPs to encode past hand waypoints to waypoint features $X^\text{wp}\in \mathbb{R}^{(N_\text{p}+\ell)\times x}$, which are further concatenated with $X^\text{vl}$ and tiled $X^\text{task}$. As showcased in Fig.~\ref{fig:fusion_module}, the additional target indicators are further concatenated with them to designate the prediction target. The target indicators consist of the tiled one-hot embeddings $X^\text{tar}\in \mathbb{R}^{e}$, where $e$ denotes the total number of prediction targets. 
Ultimately, the feature combos are fused by MLPs to produce the HM latents $F^\text{hm}_\text{p}\in \mathbb{R}^{N_\text{p}\times f}$ and $F^\text{hm}_\text{f}\in \mathbb{R}^{N_\text{f}\times f}$ for the following HMF diffusion. $f$ represents the channel dimension of HM latents. $F^\text{hm}_\text{f}$ only exists in the training stage for reconstruction supervision since noisy future latents $F^\text{hm}_\text{noise}\in \mathbb{R}^{N_\text{f}\times f}$ is concatenated to $F^\text{hm}_\text{p}$ in the inference stage of the HMF diffusion.
By seamlessly integrating the target indicators into the framework, Uni-Hand identifies HM latents of different hand centers and joints, and achieves multi-target forecasting with the following HMF diffusion.

\begin{figure}[t]
  \centering
  \includegraphics[width=1\linewidth]{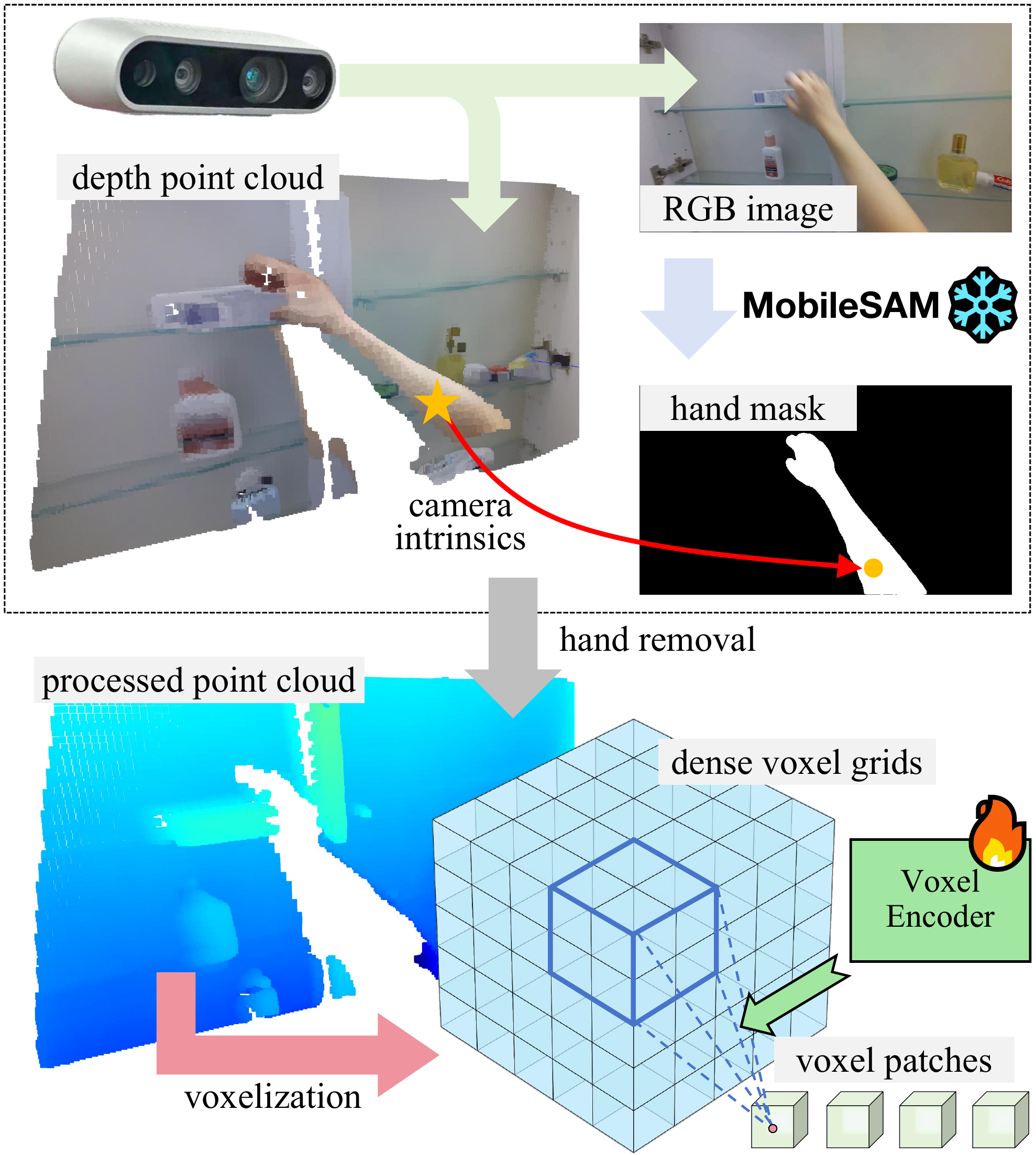}
  \caption{Hand removal for purified point clouds. We regard the voxel patches encoded by the voxel encoder as 3D global context for the denoising process in the HMF diffusion.}
  \label{fig:hand_removal}
  \vspace{-0.5cm}
\end{figure}

\subsubsection{Voxel Encoder}
\textbf{Point Cloud Preprocessing.} Uni-Hand advances 3D structure-awareness by absorbing point cloud observations for more reasonable HMF compared to the existing methods which only receive RGB images. The headset RGB-D camera captures a depth image aligned with its RGB counterpart $I_t$, which is directly transformed to the input point cloud $D_t \in O_t$ for Uni-Hand. To address the practical constraints of computational efficiency and memory requirements, we convert the holistic point cloud into a more compact voxelized form, as shown in Fig.~\ref{fig:hand_removal}. Specifically, we first exploit MobileSAM~\cite{mobile_sam} to remove depth points projected to arms for each $D_t$. This is because arm movements generate artifact points hindering global 3D representation accuracy during the following multi-frame aggregation. Subsequently, all the purified point clouds within the past egocentric observations $\mathcal{O}$ are transformed into a unified global coordinate system by visual odometry. 
Then we aggregate them into voxel grids to avoid disturbance of the unordered point cloud structure, and further reduce the data volume. 

\textbf{Voxel Encoding.} After the above point cloud preprocessing, the voxel encoder in Uni-Hand encodes the dense 3D voxels into a sparser representation $X^\text{vox}\in \mathbb{R}^{N_\text{vox}\times f}$ through 3D convolutions. That is, each input point cloud corresponds to $N_\text{vox}$ voxel patches with the same channel dimension $f$ as HM latents. 
Notably, while these voxel patches share dimensional alignment with HM latents, we do not integrate them into HM latents. Instead, they are maintained as separate global contexts of 3D interaction environments for the denoising model in the following HMF diffusion. This stems from the fact that the unified global representation of aggregated point clouds has not been time-varying compared to sequential egocentric observations. Merging it into time-varying HM latents is not required.

\subsection{Dual-Branch Diffusion Model}
\label{sec:dbdm}

\subsubsection{Hand-Head Motion Coordination}
\begin{figure}[t]
  \centering
  \includegraphics[width=1\linewidth]{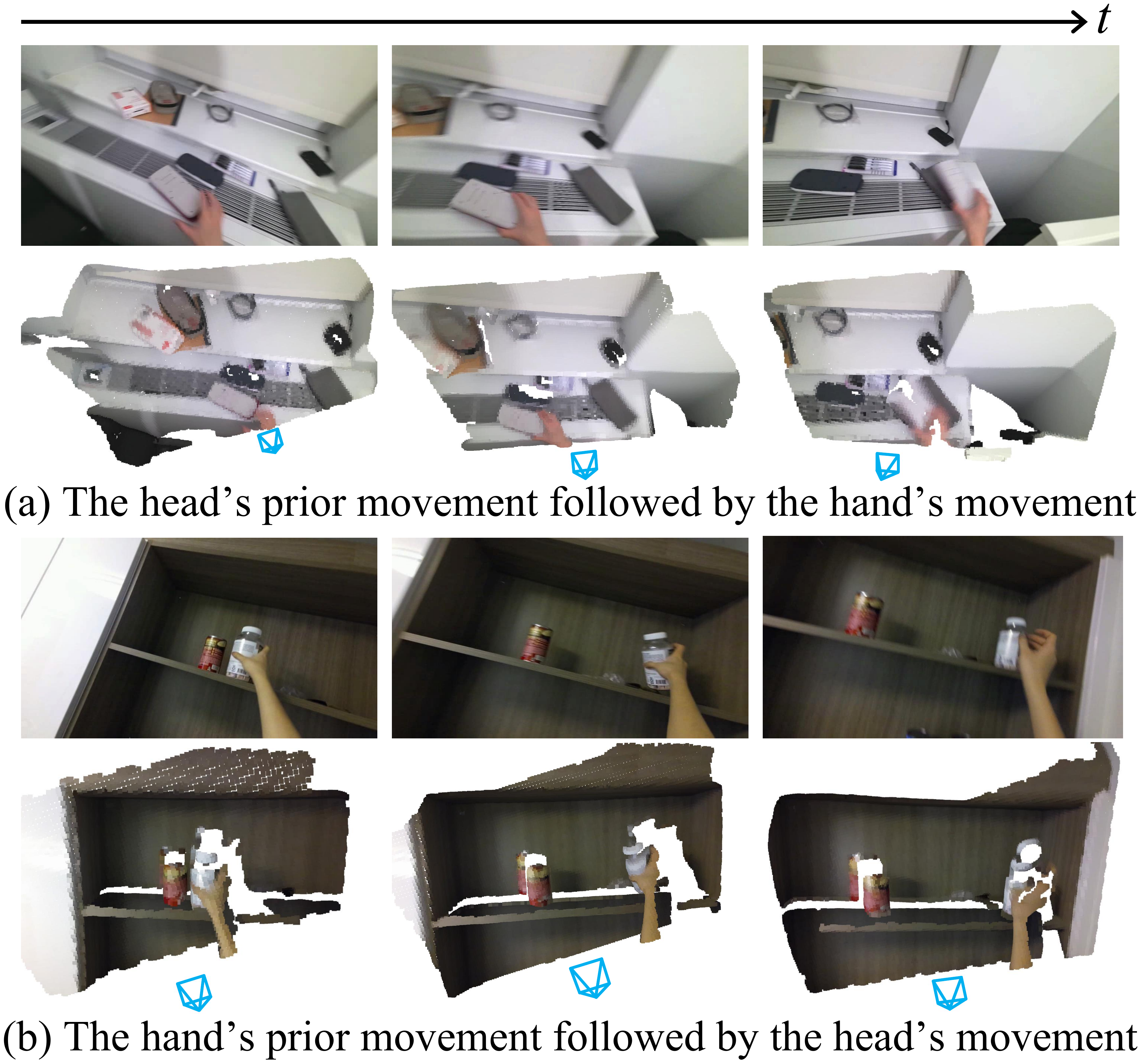}
  \caption{Examples of head movement (corresponding to camera egomotion) and hand movement entangled during the hand-object interaction process in egocentric views in the EgoPAT3D dataset~\cite{li2022egocentric}. Here we present the RGB images and point clouds, as well as camera poses to clarify the hand-head motion trend.}
  \label{fig:coupled_motion}
  \vspace{-0.5cm}
\end{figure}

Our proposed dual-brach diffusion model is aimed at capturing the synergy between hand movements and headset camera egomotion, and concurrently predicting future HM latents and EM latents. 
The hand-head motion coordination within the future interaction process is reflected in three aspects: (1) Hand movements often follow head motion, as the head’s prior motion provides key visual cues for trajectory planning (Fig.~\ref{fig:coupled_motion}(a)), (2) head movements may follow hand actions, as subconscious and faster hand motions can lead to head responses (Fig.~\ref{fig:coupled_motion}(b)), and (3) humans keep hand movements within egocentric views to ensure accurate target contact (Fig.~\ref{fig:coupled_motion}(a) and Fig.~\ref{fig:coupled_motion}(b)). We argue that predicting hand motion agnostic to future head motion does not align with real human behavior planning. Instead, explicitly decoupling the entangled hand-head movements helps prediction models to better understand synergy motion patterns and potential intentions of interaction. In light of this, we propose the dual-branch diffusion model, encompassing the EMF diffusion and HMF diffusion to predict future EM latents (headset camera egomotion) and HM latents (hand motion) concurrently.

\subsubsection{EMF Diffusion}
The EMF diffusion is developed to denoise future EM latents, thus obtaining future head motion trend for the following hand motion forecasting. As shown in Fig.~\ref{fig:system_overview}(b), the EMF diffusion leverages Mamba~\cite{gu2023mamba} as the denoising model to convert the noisy future EM latents $F^\text{em}_\text{noise}$ to $\hat{F}^\text{em}_\text{f}\in \mathbb{R}^{N_\text{f}\times f}$ conditioned on the past EM latents $F^\text{em}_\text{p}$. The predicted $\hat{F}^\text{em}_\text{f}$ implicitly represents future head motion trend during the interaction process. \myblue{Since head motion exhibits simpler dynamics than hand motion, and head motion forecasting must be sufficiently efficient to provide conditioning signals for hand motion forecasting, we adopt the vanilla Mamba blocks for efficient denoising in our proposed EMF diffusion.} Besides, we omit the process of decoding the future EM latents into specific homography matrices. This eliminates the ambiguity inherent in selecting supervision signals for alternative homography parameterizations~\cite{detone2016deep}.

\begin{figure}[t]
  \centering
  \includegraphics[width=1\linewidth]{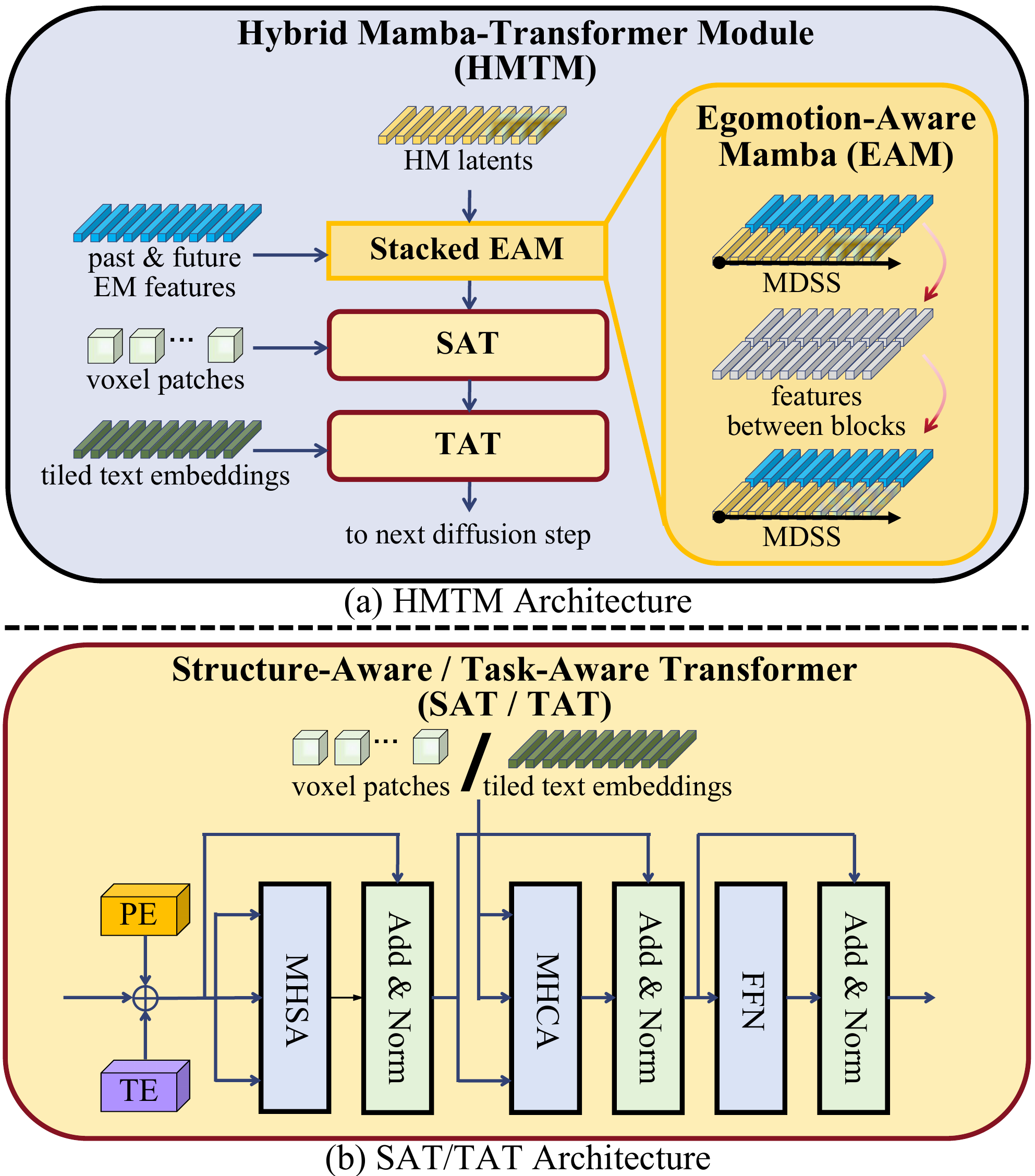}
  \vspace{-0.5cm}
  \caption{Architecture of the hybrid Mamba-Transformer module. It denoises HM latents with stacked EAM, SAT, and TAT blocks.}
  \label{fig:mamba_transformer}
  \vspace{-0.5cm}
\end{figure}

\subsubsection{HMF Diffusion}
With the predicted camera egomotion trend, the HMF diffusion aims to forecast future HM latents, which will be transformed into the explicit future hand trajectories and interaction states by the devised decoders. 
As shown in Fig.~\ref{fig:system_overview}(b), the HMF diffusion receives past HM latents $F^\text{hm}_\text{p}$ to predict future counterparts $\hat{F}^\text{hm}_\text{f}$, conditioned on $F^\text{em}_\text{p}$ and $\hat{F}^\text{em}_\text{f}$ predicted by the EMF diffusion. 
Here we propose a novel hybrid Mamba-Transformer module (HMTM) as the denoising model. As showcased in Fig.~\ref{fig:mamba_transformer}(a), HMTM comprises three key components: (1) stacked egomotion-aware Mamba (EAM) blocks adapted from MADiff~\cite{ma2024madiff}, which employ Motion-Driven Selective Scan (MDSS) to seamlessly incorporate egomotion homography features into Mamba's state transitions, (2) our newly designed structure-aware Transformer (SAT), which absorbs voxel patches as 3D global context by multi-head corss attention, and (3) task-aware Transformer (TAT), where we inject text embeddings of task instructions into the HMF denoising process.

\textbf{Stacked Egomotion-Aware Mamba (EAM).}
Concretely, we first concatenate $F^\text{em}_\text{p}$ with the predicted $\hat{F}^\text{em}_\text{f}$ to $\hat{F}^\text{em}_\text{pf} \in \mathbb{R}^{(N_\text{p}+N_\text{f})\times f}$ along the time dimension, leading to the holistic feature sequence of headset camera egomotion in past and future time horizons. Similarly, we  concatenate $F^\text{hm}_\text{p}$ with the sampled noise $F^\text{hm}_\text{noise}$ to $F^\text{hm}_\text{pf} \in \mathbb{R}^{(N_\text{p}+N_\text{f})\times f}$. Then we implement MDSS in EAM for each denoising step of the HMF diffusion, to denoise the future part of $F^\text{hm}_\text{pf}$ conditioned on the holistic sequential EM latents $\hat{F}^\text{em}_\text{pf}$. 
\myblue{Therefore, we integrate egomotion-aware Mamba's powerful ability of temporal feature modeling, as well as head motion awareness into Uni-Hand. It can narrow the visual gap between the current camera view and the prediction view, and adaptively retain critical causality across temporal sequences, as demonstrated in the previous work~\cite{ma2024madiff}.}

\textbf{Structure-Aware Transformer (SAT).}
The output of the stacked EMA blocks is further processed by the structure-aware Transformer in HMTM. SAT explicitly absorbs 3D global contexts of interaction environments for each denoising step of the HMF diffusion.
As illustrated in Fig.~\ref{fig:mamba_transformer}(b), after the standard positional/temporal encoding (PE/TE)~\cite{gong2022diffuseq,ma2024diff}, we sequentially implement multi-head self-attention (MHSA) on HM latents $F^\text{hm}_\text{pf}$, and multi-head cross-attention (MHCA) between the output of MHSA and voxel patches $X^\text{vox}$. Specifically, for MHCA input, we use the HM latents from the preceding MHSA and Add\&Norm as the query, and let the voxel patches $X^\text{vox}$ be the key and value.
Inspired by human perception of 3D spatial relationships with stereo vision, Uni-Hand incorporates voxel patches as environmental global context through MHCA in each denoising step. This enables more plausible hand motion forecasting thanks to the potential layout understanding and collision avoidance from 3D structure awareness.

\begin{figure}[t]
  \centering
  \includegraphics[width=1\linewidth]{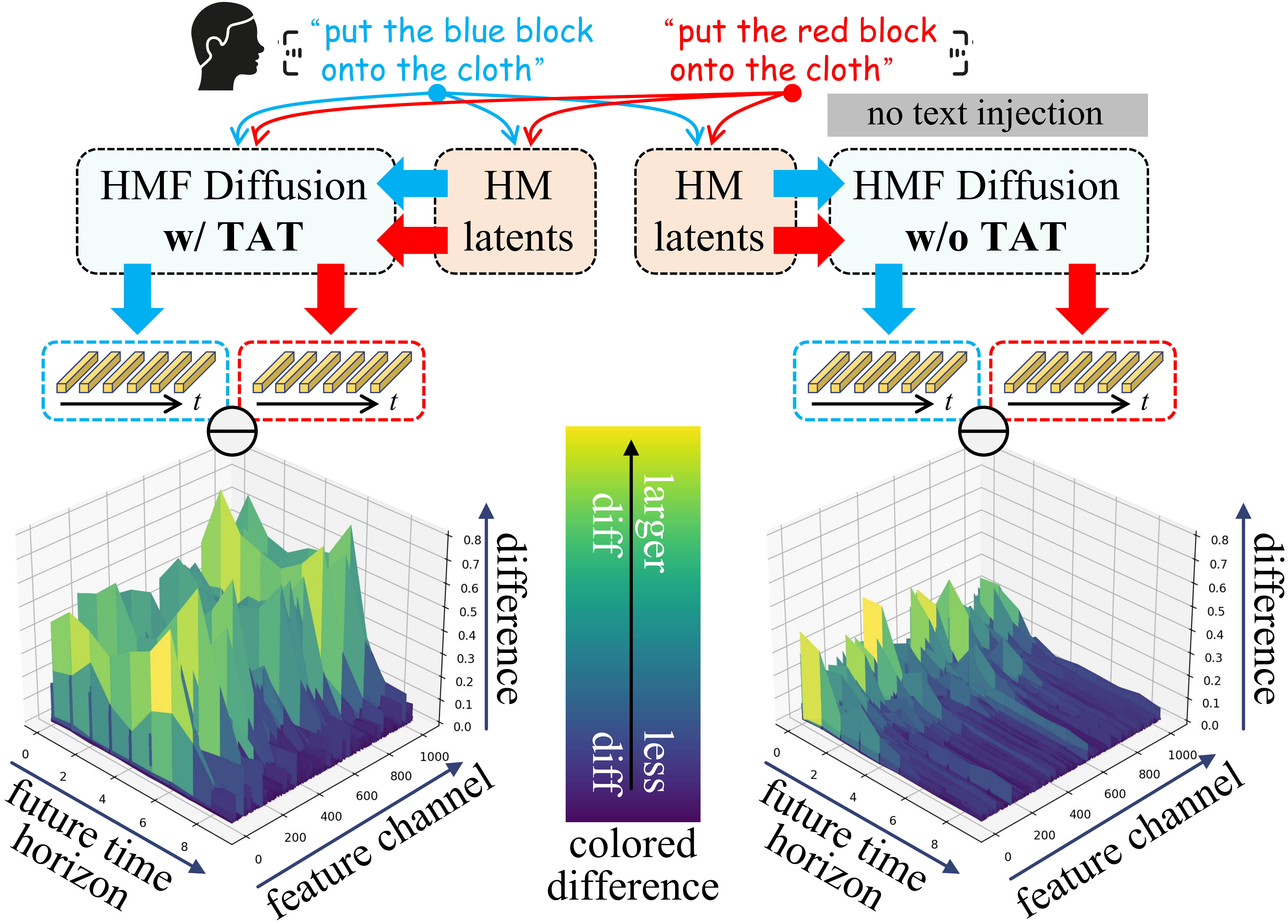}
  \caption{Difference matrices of denoised future HM latents between two text instructions, ``put the blue block onto the square cloth'' and ``put the red block onto the square cloth''. They are visualized as height fields. This illustration reflects the effectiveness of our devised TAT to enhance the awareness of specific tasks in hand feature prediction.}
  \label{fig:text_diff}
  \vspace{-0.6cm}
\end{figure}

\textbf{Task-Aware Transformer (TAT).}
Following the devised SAT, the task-aware Transformer injects specific task information into hand motion forecasting. It facilitates the model's awareness of specific human intentions or downstream task instructions.
As shown in Fig.~\ref{fig:mamba_transformer}(b), TAT has the same architecture as SAT. However, it performs multi-head cross-attention between the output of its MHSA and the tiled text embeddings $X^\text{task}$ of the task instruction. We found that only merging the task-aware text embeddings into HM latents (introduced in Sec.~\ref{sec:vfm}) is not enough to facilitate generating future hand waypoints that comply with the input task instruction. The reason could be that the task specificity inherent in HM latents is continuously weakened as the network depth increases and progressive denoising operations are performed in the HMF diffusion. We thus need to continuously inject original task information by MHCA to enhance the model's task discrimination capability. Fig.~\ref{fig:text_diff} illustrates the different matrices of denoised future HM latents
between two text instructions, ``put the blue block onto the square cloth'' and ``put the red block onto the
square cloth''. They are computed by $|\hat{F}^\text{hm,blue}_\text{f}-\hat{F}^\text{hm,red}_\text{f}|$ to present the feature difference from the two generally similar text instructions with different key targets (blue block vs. red block). As can be seen, after introducing the task-aware Transformer, besides only fusing text embeddings into HM latents, the feature discrepancies between the predicted future HM latents are significantly pronounced. That is, TAT helps the prediction model to distinguish different interaction tasks for more reasonable hand motion forecasting. Its positive effect becomes more obvious in later timings, which will directly decide how well the predicted hand motion interacts with the target object. We will further ablate the task-aware text embedding injection in Sec.~\ref{sec:exp_albation}. Note that we deploy TAT in Uni-Hand only when the task being performed by human hands is known to us. The output latents of SAT will be directly fed to the next diffusion step in task-agnostic HMF.

After the last denoising diffusion step in the HMF diffusion, we obtain the denoised HM latents $\hat{F}^\text{hm}_\text{pf}$.
As shown in Fig.~\ref{fig:system_overview}(b), the future part $\hat{F}^\text{hm}_\text{f}\in \mathbb{R}^{N_\text{f}\times f}$ of $\hat{F}^\text{hm}_\text{pf}$ is ultimately decoded by the MLP-based hand trajectory decoder and interaction state decoder to future 2D/3D hand waypoints $\hat{\mathcal{H}}_\text{f}=\{\hat{H}_t\}_{t=1}^{N_\text{f}}$ and future interaction states $\hat{\mathcal{G}}_\text{f}=\{\hat{G}_t\}_{t=1}^{N_\text{f}}$. The denoised HM features, predicted waypoints and interaction states, can also be regarded as affordances for multiple downstream tasks, which will be validated in Sec.~\ref{sec:eval_downstream}.

\vspace{-0.3cm}

\subsection{Training and Inference}
\label{sec:training_and_inf}

\subsubsection{Partial Noising and Denoising}
We exploit partial noising/denoising proposed by Gong~\etal~\cite{gong2022diffuseq} for the training and inference stages of both EMF diffusion and HMF diffusion. Specifically, we anchor the past latents $F^\text{em}_\text{p}$ and $F^\text{hm}_\text{p}$ in forward and reverse steps. In practice, although the EM latents and HM latents from the past time horizons are modified by noise corruption and denoising operations, they are manually overridden with their initial values (obtained via Multi-Modal Feature Extraction) after each diffusion step.

\subsubsection{Loss Functions}
We train Uni-Hand end-to-end, using six losses: (1) $\mathcal{L}^\text{em}_\text{VLB}$ for recovering future EM latents in the EMF diffusion, (2) $\mathcal{L}^\text{hm}_\text{VLB}$ for recovering future HM latents in the HMF diffusion, (3) trajectory displacement loss $\mathcal{L}_\text{dis}$, (4) trajectory angle loss $\mathcal{L}_\text{angle}$, (5) regularization term $\mathcal{L}_\text{reg}$, and (6) interaction state loss $\mathcal{L}_\text{state}$. We detail these losses as follows.

\textbf{Losses for EMF Diffusion.}
To generate reasonable motion trends of the headset camera by the EMF diffusion, we set $\mathcal{L}^\text{em}_\text{VLB}$ for recovering future EM latents following the diffusion-related loss used by~\cite{ma2024diff}:
\begin{align}
\mathcal{L}^\text{em}_\text{VLB} &= \sum_{s=2}^S||\mathbfz^\text{em}_0-f_{\scriptscriptstyle\text{EMF}}(\mathbfz_s^\text{em},s)||^2 + ||F^\text{em}_\text{f}-\hat{F}^\text{em}_\text{f}||^2,
\label{eq:em_loss}
\end{align}
where $s$ is the index of each diffusion step, and $S$ is the total diffusion steps. $\mathbfz^\text{em}_s$ is the latent after the $s$-th denoising step. $f_{\scriptscriptstyle\text{EMF}}$ denotes the function of the denoising model in the EMF diffusion. $F^\text{em}_\text{f}$ is the GT future EM latents from the observations in future time horizons, and $\hat{F}^\text{em}_\text{f}$ is the counterparts finally predicted by the EMF diffusion as mentioned before. 

\textbf{Losses for HMF Diffusion.} 
The diffusion-related loss $\mathcal{L}^\text{hm}_\text{VLB}$ supervising the predicted future HM latents in the HMF diffusion is similarly defined as:
\begin{align}
\mathcal{L}^\text{hm}_\text{VLB} &= \sum_{s=2}^S||\mathbfz^\text{hm}_0-f_{\scriptscriptstyle\text{HMF}}(\mathbfz_s^\text{hm},s)||^2 + ||F^\text{hm}_\text{f}-\hat{F}^\text{hm}_\text{f}||^2,
\label{eq:hm_loss}
\end{align}
where $\mathbfz^\text{hm}_s$ indicates the latent after the $s$-th denoising step. $f_{\scriptscriptstyle\text{HMF}}$ represents the function of the denoising model in the HMF diffusion. $F^\text{hm}_\text{f}$ is the GT future HM latents encoded from the observations in future time horizons. $\hat{F}^\text{hm}_\text{f}$ is the counterparts predicted by the HMF diffusion. 

In addition, we utilize the trajectory displacement loss $\mathcal{L}_\text{dis}$ and the trajectory angle loss $\mathcal{L}_\text{angle}$ to optimize the hand waypoints decoded by the hand trajectory decoder:
\begin{align}
\mathcal{L}_\text{dis}&=\frac{1}{N_\text{f}}\sum_{t=1}^{N_\text{f}}D_\text{dis}(H_t,\hat{H}_t), \\
\mathcal{L}_\text{angle}&=\frac{1}{N_\text{f}}\sum_{t=0}^{N_\text{f}-1}D_\text{cos}(H_{t+1}-H_t,\hat{H}_{t+1}-\hat{H}_{t}),
\label{eq:dis_ang_loss}
\end{align}
where $D_\text{dis}(\cdot)$ denotes the Euclidean distance between predicted hand waypoints and GT ones. $\hat{H}_t$ is the output hand waypoints of the hand trajectory decoder, while ${H}_t$ is its GT counterpart. $D_\text{cos}(\cdot)$ represents the cosine similarity between two input motion vectors.

Following the previous works~\cite{ma2024diff,ma2024madiff}, we also introduce the regularization term $\mathcal{L}_\text{reg}$ to penalize the difference between the pseudo trajectory predictions $\tilde{H}_t$ and $H_t$:
\begin{align}
\mathcal{L}_\text{reg}&=\frac{1}{N_\text{f}}\sum_{t=1}^{N_\text{f}}D_\text{dis}(H_t,\tilde{H}_{t}),
\label{eq:reg_loss}
\end{align}
where $\tilde{H}_t$ is obtained by directly feeding $F^\text{hm}_\text{f}$ to the hand trajectory decoder rather than the predicted $\hat{F}^\text{hm}_\text{f}$. This regularization helps to improves training stability and optimization performance.

To optimize interaction state forecasting, we utilize the interaction state loss $\mathcal{L}_\text{int}$ defined as:
\begin{align}
\mathcal{L}_\text{int}=\text{CE}(\mathcal{G}_\text{f},\hat{\mathcal{G}}_\text{f}),
\label{eq:inter_loss}
\end{align}
where $\text{CE}(\cdot)$ is the binary cross-entropy loss function. $\hat{\mathcal{G}}_\text{f}$ is the predicted interaction states decoded by the interaction state decoder, and $\mathcal{G}_\text{f}$ is its GT label. Through the supervision of the interaction state loss, the timings of hand-object contact will be predicted with probabilities closer to 1.

\textbf{Total Loss Function.} The total loss function to train our proposed Uni-Hand is the weighted sum of all the above losses, which is expressed as:
\begin{align}
\mathcal{L}_\text{total}=\lambda^\text{em}_\text{VLB}\mathcal{L}^\text{em}_\text{VLB} + \lambda^\text{hm}_\text{VLB}\mathcal{L}^\text{hm}_\text{VLB} + \lambda_\text{dis}\mathcal{L}_\text{dis} + \lambda_\text{angle}\mathcal{L}_\text{angle} \nonumber\\  + \lambda_\text{reg}\mathcal{L}_\text{reg} + \lambda_\text{int}\mathcal{L}_\text{int},
\label{eq:total_loss}
\end{align}
where $\lambda^\text{em}_\text{VLB}$, $\lambda^\text{hm}_\text{VLB}$, $\lambda_\text{dis}$, $\lambda_\text{angle}$, $\lambda_\text{reg}$, and $\lambda_\text{int}$ are weights to balance the trade-off between these losses.

\begin{figure*}[t]
  \centering
  \includegraphics[width=0.9\linewidth]{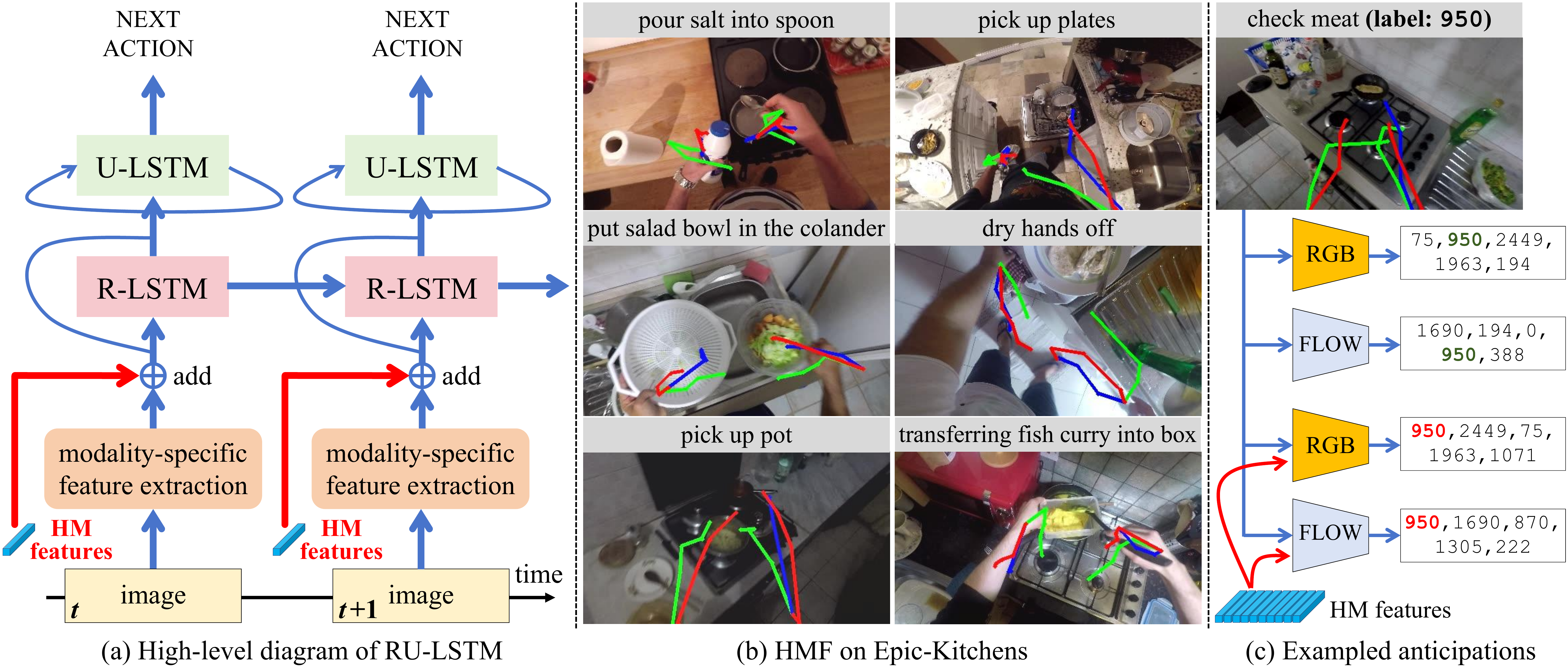}
  \caption{Illustration of testing Uni-Hand in the downstream action anticipation task. In the egocentric images of (b) and (c), the red, green, and blue lines denote hand center waypoints predicted by Uni-Hand, OCT~\cite{liu2022joint}, and GT labels, respectively.}
  \label{fig:action_anticipation}
  \vspace{-0.5cm}
\end{figure*}

\section{Experimental Results}
\label{sec:exp}

\subsection{Benchmarks}
\myblue{We evaluate our proposed Uni-Hand and the existing baselines on five publicly available datasets including EgoPAT3D-DT~\cite{li2022egocentric,bao2023uncertainty}, H2O-PT~\cite{kwon2021h2o,bao2023uncertainty}, HOT3D-Clips~\cite{banerjee2024hot3d}, Ego4D~\cite{grauman2022ego4d}, and Epic-Kitchens-55~\cite{damen2018scaling}, and two self-recorded datasets including the Cup-Apple-Box-Hand (CABH) benchmark and Hand-ALOHA-Transfer (HAT) benchmark.} In the HAT benchmark, we conduct human-robot transfer experiments on a real robot platform, ALOHA~\cite{zhao2023learning}, to present how
Uni-Hand effectively supports manipulation tasks in real-world setups.
\subsubsection{Public Datasets}

We follow the setups of the prior works~\cite{bao2023uncertainty,ma2025mmtwin} to organize EgoPAT3D-DT, H2O-PT, and HOT3D-Clips datasets. Please refer to Supp. Mat., Sec.~A for more details. \myblue{For Ego4D, we attend to the 5-frame annotated clip with unconstrained head motions for each prediction in its future hand prediction benchmark, and let all the evaluated approaches receive the first three frames (three frames preceding the pre-condition frame) as input to predict the hand locations in the subsequent two frames (pre-condition frame and contact frame).}
To further demonstrate how our Uni-Hand supports the downstream action anticipation, early action recognition, and action recognition tasks, we organize the Epic-Kitchens-55 dataset following the splits of OCT~\cite{liu2022joint}. 
The evaluation paradigm for the downstream action anticipation task is illustrated in Fig.~\ref{fig:action_anticipation}. As shown in Fig.~\ref{fig:action_anticipation}(a), we select the widely-used and off-the-shelf action anticipation framework, RU-LSTM~\cite{furnari2020rolling}, to present the feature enhancement from our HMF method. Concretely, we first train Uni-Hand with annotated hand center waypoints in Epic-Kitchens-55, leading to a good HMF capability as depicted in Fig.~\ref{fig:action_anticipation}(b). Afterwards, when implementing downstream action anticipation, we directly add the HM features (latents recovered by HMF diffusion of Uni-Hand) to the modality-specific features from the vanilla RU-LSTM. The combined features are further processed by the following rolling-unrolling operations through R-LSTM and U-LSTM. Then we optimize RU-LSTM for the action anticipation task, by minimizing the loss between the predicted action categories (linearly transformed from U-LSTM outputs) and the ground-truth labels. Note that we consider the modality-specific branches of RU-LSTM, i.e., the RGB branch and the flow branch, to demonstrate that our HMF method can enhance both RGB and flow features for higher anticipation accuracy. We do not use the object branch of RU-LSTM due to the mismatch in channel dimensions between HMF latents and object features. The feature enhancement operations for early action recognition and action recognition tasks are similar to action anticipation, while we only consider the ANTICIPATION period of RU-LSTM and discard its ENCODING period. Please refer more details about RU-LSTM structures to the original paper~\cite{furnari2020rolling}.

\begin{figure}[t]
  \centering
  \includegraphics[width=1\linewidth]{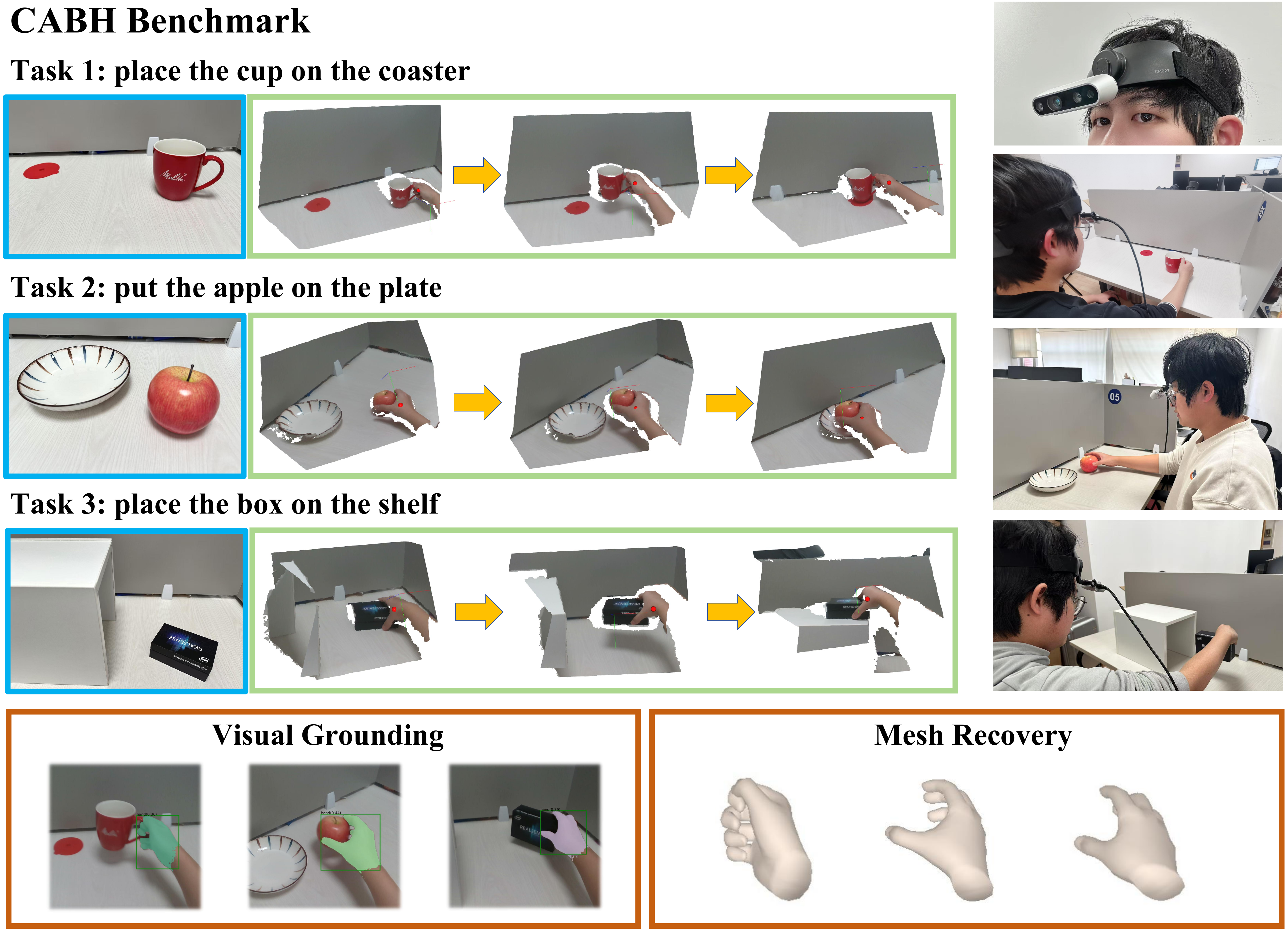}
  \caption{Our self-collected CABH benchmark includes three hand-object interaction tasks to efficiently evaluate HMF.}
  \label{fig:cabh_benchmark}
  \vspace{-0.5cm}
\end{figure}

\subsubsection{Cup-Apple-Box-Hand Benchmark}
\label{sec:cabh_bench}
To further demonstrate that our method has the potential to scale up with low-cost devices for data collection, we used headset RealSense D435i to collect 1200 egocentric videos for three real-world tasks, i.e., \textit{place the cup on the coaster} (Task~1), \textit{put the apple on the plate} (Task~2), and \textit{place the box on the shelf} (Task~3), to build the CABH benchmark (see Fig.~\ref{fig:cabh_benchmark}). For each task, 350 video clips are used for training with the other 50 clips for evaluation. Each clip has around 5 seconds, with the first 50\% regarded as the past sequences and the latter 50\% as the future ones. To validate that Uni-Hand can achieve plausible hand motion forecasting using only egocentric vision without expensive MoCap systems for GT annotations, we utilize the visual grounding model, Grounded SAM~\cite{ren2024grounded}, and the hand mesh recovery approach, HaMeR~\cite{pavlakos2024reconstructing}, to label 2D hand centers and joints on videos. Their 3D positions can be further recovered with depth observations of the RGB-D camera. To demonstrate that the choice of prediction canvas does not constrain the superiority of Uni-Hand, we directly predict future hand waypoints with each future frame as the canvas.

\subsubsection{Hand-ALOHA-Transfer Benchmark}
\label{sec:hat_bench}
\begin{figure}[t]
  \centering
  \includegraphics[width=1\linewidth]{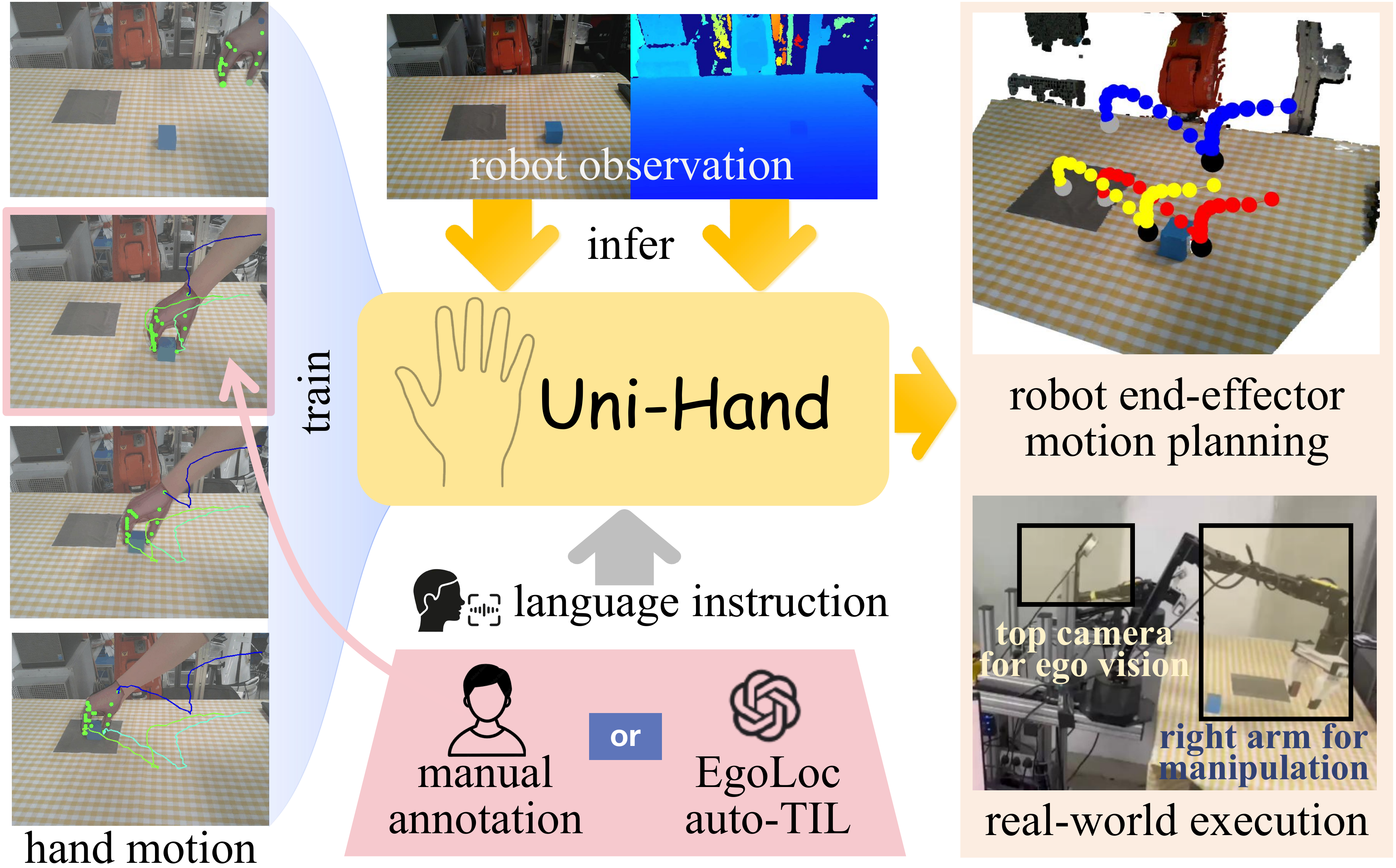}
  \caption{Our scheme to deploy Uni-Hand to real-world robotic manipulation tasks.}
  \label{fig:deployment_scheme}
  \vspace{-0.5cm}
\end{figure}

\textbf{Task Definition.} We also propose the Hand-ALOHA-Transfer benchmark by recording hand motion with the top camera mounted on ALOHA~\cite{zhao2023learning}, as shown in Fig.~\ref{fig:deployment_scheme}. The top camera, RealSense LiDAR Camera L515, captures egocentric RGB images and point clouds (depth) for hand motion forecasting and end-effector action prediction. In the HAT benchmark, we introduce \textbf{five} manipulation tasks: 
\begin{itemize}[leftmargin=1em]
\item \textbf{Push Task} (\taskAA): push the block onto the cloth. 400 hand motion videos in total for training Uni-Hand. 10 trials to evaluate real-world performance.
\item \textbf{Pick-and-Place Task} (\taskBB): put the block onto the cloth. 400 hand motion videos in total for training Uni-Hand. 10 trials to evaluate real-world performance.
\item \textbf{Harder Pick-and-Place Task} (\taskCC): stack the blue block onto the red block. 400 hand motion videos in total for training Uni-Hand. 10 trials to evaluate real-world performance.
\item \textbf{Language-Conditioned Pick-and-Place Task} (\taskDD): put the \{\textit{blue}, \textit{red}\} block onto the cloth. 800 hand motion videos for training, 400 videos per category (\textit{blue}, \textit{red}). 10 trials to evaluate real-world performance, 5 trials per category.
\item \textbf{Long-Horizon Pick-and-Place Task} (\taskEE): put the block, the toy, and the banana onto the plate. 800 hand motion videos in total for training Uni-Hand. 10 trials to evaluate real-world performance.
\end{itemize}
These multiple manipulation tasks are used to evaluate both HMF performance and human-robot policy transfer ability.

\textbf{Deployment Scheme.} As illustrated in Fig.~\ref{fig:deployment_scheme}, we train Uni-Hand with recorded hand motion data, enabling it to generate complete executable actions for the robot end-effector in a look-then-move manner. Concretely, the predicted 3D trajectories are used for tracking, and the predicted interaction states help the robot to decide when to open and close its gripper. 
The GT 3D hand waypoints are captured by HaMeR~\cite{pavlakos2024reconstructing}. To label the GT interaction states, we introduce two approaches: (1) manual annotation, and (2) automated temporal interaction localization (auto-TIL) using EgoLoc~\cite{zhang2025egoloc}.
\myblue{In both the training and inference stages with the HAT benchmark, only the first observation frame (i.e., $N_\text{p}=1$, agnostic to hands) is used as the egocentric visual inputs. The prediction model directly predicts the global motion with complete waypoints and interaction states, instead of step-by-step reasoning conditioned on iterative observations and states. This is because the robot manipulates autonomously without visual access to human hands, depending entirely on environmental observations. We should not encode observations involving human hands in the training stage. Our HAT benchmark thus evaluates the HMF methods' ability to generate the holistic trajectory and interaction states from scratch without the constraints of observed motion (See Supp. Mat., Sec.~S for more details).} Besides, for all the robotic manipulation tasks, we curate the mapping from human hand grasp poses to robot grasp poses, which is detailed in Supp. Mat., Sec.~C.

\vspace{-0.3cm}
\subsection{Uni-Hand Configurations}
\label{sec:unihand_config}
For vision-language features from images and universal text prompt shown in Fig.~\ref{fig:fusion_module}, we extract the outputs of the deepest cross-modality multi-head attention module (X-MHA) in GLIP~\cite{Li_2022_CVPR}, which are further processed by the channel adapter proposed by~\cite{ma2024madiff}. Besides, in Fig.~\ref{fig:fusion_module}, the channel dimension of all the features before concatenation is set as $1024$, and the counterpart of the target indicators is determined by the number of prediction targets (e.g., 5 if predicting five hand joints). To generate 3D global context shown in Fig.~\ref{fig:hand_removal}, we convert input point clouds into voxel grids of dimensions $20\times 20\times 20$ with a resolution of $0.05$\,m. Each voxel patch has a size of $27 \times 1024$. For both EMF and HMF diffusion, we set the channel dimension of the latent features to $1024$. The total number of diffusion steps is set to $1000$, while the EMF diffusion takes only one step to predict EM latents for high efficiency, and the HMF diffusion takes 100 steps to predict future HM latents. We adaptively adjust the input and output dimension of the MLPs for encoding and decoding hand trajectories according to the predictive space (2D/3D). 
Notably, for hand-ALOHA motion transfer, the camera homography is an identity matrix since the robot base is fixed in our HAT benchmark. Besides, we empirically found that the HMF diffusion can generate more stable trajectories for the robot end-effector when increasing the denoising steps to 1000. There is a trade-off between inference efficiency and planning accuracy.
\myblue{The hybrid pattern of Mamba and Transformer in our devised hybrid Mamba-Transformer module is designed according to the ablation study in Supp. Mat., Sec.~E.} We train Uni-Hand using AdamW optimizer \cite{kingma2014adam} with a learning rate of 5e-5 for 1K epochs on EgoPAT3D-DT, H2O-PT, Ego4D, and our self-recorded benchmarks, and with a learning rate of 5e-6 for 2K epochs on the HOT3D-Clips dataset.

\begin{table*}[t]
\small
\setlength{\tabcolsep}{12pt}
\center
\renewcommand\arraystretch{0.7}
\caption{\myblue{Comparison of performance on 3D hand trajectory prediction on the publicly available datasets. Best and secondary results are viewed in \textbf{bold black} and \mygreen{green} colors respectively.}}
\vspace{-0.2cm}
\begin{tabular}{l|cc|cc|cc|cc}
\toprule
\multicolumn{1}{l|}{\multirow{2}{*}{Approach}}   & \multicolumn{2}{c|}{EgoPAT3D-DT (seen)} & \multicolumn{2}{c|}{EgoPAT3D-DT (unseen)}  & \multicolumn{2}{c|}{H2O-PT} & \multicolumn{2}{c}{HOT3D-Clips} \\ \cmidrule{2-9} 
\multicolumn{1}{c|}{}                                                                               & ADE\,$\downarrow$    & FDE\,$\downarrow$ & ADE\,$\downarrow$   & FDE\,$\downarrow$ & ADE\,$\downarrow$   & FDE\,$\downarrow$ & ADE\,$\downarrow$   & FDE\,$\downarrow$   \\ \cmidrule{1-9}                 
CVH \cite{ma2024diff}  & 1.100  	& 1.278  & 0.952   & 1.018  & 0.146    & 0.148 & 1.273 & 1.358 \\
OCT$^{*}$ \cite{liu2022joint}   &0.370    & 0.524  &  0.309   & 0.397  & 0.103  &0.126 & 0.188  & 0.215 \\ 
USST$^*$ \cite{bao2023uncertainty}  & \mygreen{0.183}	  & \mygreen{0.341}	 & \mygreen{0.120}  & \textbf{0.185}	&\mygreen{0.031}	 & \mygreen{0.052}  & 0.123   &  0.157  \\ 
S-Mamba \cite{wang2024mamba} & 0.185    & 0.355	&0.138	&0.207	 & 0.038 	& 0.074  & \mygreen{0.117}	 &  \mygreen{0.132} \\ 
Diff-IP3D \cite{ma2024diff}  & 0.199     &0.377	 &0.156  &0.229	  &0.049    & 0.081   & 0.147 	& 0.164 \\ 
\myblue{EMAG3D} \cite{hatano2024emag}  & \myblue{0.207}  & \myblue{0.365}   & \myblue{0.166}  & \myblue{0.209}  & \myblue{0.062}  & \myblue{0.085} & \myblue{0.141}  & \myblue{0.158} \\
\myblue{EER3D} \cite{ma2025eer}  & \myblue{0.195}  & \myblue{0.377}   & \myblue{0.158}  & \myblue{0.233}  & \myblue{0.047}  & \myblue{0.075} & \myblue{0.150}  & \myblue{0.162} \\
MADiff3D \cite{ma2024madiff}  & \mygreen{0.183}     &0.363	 &0.139  &0.224  &0.032    & 0.059   & 0.120  & 0.147	 \\
\rowcolor{lightgray}
Uni-Hand (ours)  &\textbf{0.170}	  &\textbf{0.336}	  &\textbf{0.118}	  &\mygreen{0.189}  	&\textbf{0.030} 	&\textbf{0.050}  & \textbf{0.104}	& \textbf{0.131}\\ \bottomrule
\end{tabular}
\label{tab:compare_hand_egopat_h2o_3d}
\\ \vspace{-0.15cm}
\begin{flushleft}
\scriptsize
$^*$\,The baselines are re-evaluated according to the erratum: \url{https://github.com/oppo-us-research/USST/commit/beebdb963a702b08de3a4cf8d1ac9924b544abc4}.
\end{flushleft}
\vspace{-0.3cm}
\end{table*}

\begin{table*}[t]
\small
\setlength{\tabcolsep}{12pt}
\center
\renewcommand\arraystretch{0.7}
\caption{\myblue{Comparison of performance on 2D hand trajectory prediction on the publicly available datasets. Best and secondary results are viewed in \textbf{bold black} and \mygreen{green} colors respectively.}}
\vspace{-0.2cm}
\begin{tabular}{l|cc|cc|cc|cc}
\toprule
\multicolumn{1}{l|}{\multirow{2}{*}{Approach}}   & \multicolumn{2}{c|}{EgoPAT3D-DT (seen)} & \multicolumn{2}{c|}{EgoPAT3D-DT (unseen)}  & \multicolumn{2}{c|}{H2O-PT} & \multicolumn{2}{c}{HOT3D-Clips} \\ \cmidrule{2-9} 
\multicolumn{1}{c|}{}                                                                               & ADE\,$\downarrow$    & FDE\,$\downarrow$ & ADE\,$\downarrow$   & FDE\,$\downarrow$ & ADE\,$\downarrow$   & FDE\,$\downarrow$ & ADE\,$\downarrow$   & FDE\,$\downarrow$   \\ \cmidrule{1-9}                 
CVH \cite{ma2024diff}  & 0.180   & 0.230   & 0.188    & 0.221   & 0.206     & 0.208 & 0.437 & 0.444 \\
OCT$^{*}$ \cite{liu2022joint}   & 0.108     & 0.122   &  0.091      &  0.147   & 0.387     & 0.381  & 0.190 & 0.240  \\ 
USST$^*$ \cite{bao2023uncertainty}  & {0.082}    & {0.118}   & {0.060}       & {0.087}    & {0.040}     & \mygreen{0.068}   & 0.133 & 0.172   \\ 
S-Mamba \cite{wang2024mamba} & 0.084    & 0.141  	& 0.071	& 0.118  & 0.051  & 0.094  & 0.130   & \mygreen{0.160}  \\ 
Diff-IP2D \cite{ma2024diff}   & {0.080}    & 0.130   & {0.066}     & {0.087}    & {0.042}       & {0.074}  
 & 0.169 & 0.208   \\ 
\myblue{EMAG} \cite{hatano2024emag}  & \myblue{0.095}  & \myblue{0.143}   & \myblue{0.092}  & \myblue{0.139}  & \myblue{0.054}  & \myblue{0.093} & \myblue{0.155}  & \myblue{0.194} \\
\myblue{EER} \cite{ma2025eer}  & \myblue{0.072}  & \myblue{0.121}   & \myblue{0.056}  & \textbf{\myblue{0.080}}  & \myblue{0.039}  & \myblue{0.069} & \myblue{0.166}  & \myblue{0.204} \\
MADiff \cite{ma2024madiff}  & \mygreen{0.065}       & \textbf{0.105}     &  \mygreen{0.054}     & \mygreen{0.086}    & \mygreen{0.039}   & \mygreen{0.068} & \mygreen{0.124}  & \mygreen{0.160}   \\
\rowcolor{lightgray}
Uni-Hand (ours)  & \textbf{0.064}  & \mygreen{0.109}  & \textbf{0.052}  & {0.087}  & \textbf{0.036} & \textbf{0.065}	& \textbf{0.118} & \textbf{0.156} \\ \bottomrule
\end{tabular}
\label{tab:compare_hand_egopat_h2o_2d}
\vspace{-0.3cm}
\end{table*}

\begin{table}[t] 
\small
\setlength{\tabcolsep}{23pt}
\center
\renewcommand\arraystretch{0.7}
\caption{\myblue{Comparison of performance on hand trajectory prediction on the Ego4D dataset in 2D space. Best and secondary results are viewed in \textbf{bold black} and \mygreen{green} colors.}}
\vspace{-0.2cm}
\begin{tabular}{l|cc}
\toprule
Approach   & ADE\,$\downarrow$    & FDE\,$\downarrow$   \\ \cmidrule{1-3}    
USST~\cite{bao2023uncertainty}    & 0.196    & 0.208            \\
EMAG~\cite{hatano2024emag}    & 0.221    & 0.228            \\
Diff-IP2D~\cite{ma2024diff}    & 0.187  & 0.208             \\
S-Mamba~\cite{wang2024mamba}    & \mygreen{0.182}   & \mygreen{0.201}       \\
\rowcolor{lightgray}
Uni-Hand (ours)    & \textbf{0.176}  & \textbf{0.193}    \\ \bottomrule
\end{tabular}
\label{tab:ego4d}
\vspace{-0.2cm}
\end{table}

\begin{table}[t]
\small
\setlength{\tabcolsep}{2.5pt}
\center
\renewcommand\arraystretch{0.7}
\caption{\myblue{Comparison of performance on hand trajectory prediction on CABH in 3D and 2D spaces with automated annotations. Best and secondary results are viewed in \textbf{bold black} and \mygreen{green} colors.}}
\begin{tabular}{l|cc|cc|cc}
\toprule
\multicolumn{1}{l|}{\multirow{3}{*}{Approach}}   & \multicolumn{2}{c|}{Task 1} & \multicolumn{2}{c|}{Task 2}  & \multicolumn{2}{c}{Task 3}   \\ \cmidrule{2-7} 
\multicolumn{1}{c|}{}                                                                              &  \multicolumn{6}{c}{3D metrics}    \\ \cmidrule{1-7}    
\multicolumn{1}{c|}{}                                                                               & ADE\,$\downarrow$    & FDE\,$\downarrow$ & ADE\,$\downarrow$    & FDE\,$\downarrow$  & ADE\,$\downarrow$    & FDE\,$\downarrow$    \\ \cmidrule{1-7}    
USST~\cite{bao2023uncertainty}   & 0.102     &  0.125   &  0.109     &  0.128      & 0.103    & 0.130      \\
Diff-IP3D~\cite{ma2024diff}   & 0.053     &  0.057   &  0.055     &  0.078      & 0.064    & 0.080      \\
S-Mamba~\cite{wang2024mamba}   & \mygreen{0.045}   & \mygreen{0.055}     & \mygreen{0.050}     & \mygreen{0.072}   & \mygreen{0.058} & \mygreen{0.061}           \\
\rowcolor{lightgray}
Uni-Hand (ours) & \textbf{0.041}	& \textbf{0.052}  & \textbf{0.044}   & \textbf{0.061}   & \textbf{0.047}  & \textbf{0.053}
  \\  \cmidrule{1-7} 
  \multicolumn{1}{c|}{}                                                                              &  \multicolumn{6}{c}{2D metrics}    \\ \cmidrule{1-7}    
USST~\cite{bao2023uncertainty}   & 0.115  &  0.155   &  0.127     &  0.168      &  0.124      & 0.161      \\
Diff-IP2D~\cite{ma2024diff}   & 0.084     &  \mygreen{0.100}   &  0.094     &  0.117      &  0.109    & 0.126  \\
S-Mamba~\cite{wang2024mamba}   & \mygreen{0.080}     & 0.102     & \mygreen{0.081}     & \mygreen{0.107}    & \mygreen{0.102} & \mygreen{0.108}           \\
\rowcolor{lightgray}
Uni-Hand (ours) & \textbf{0.066}	& \textbf{0.086}  & \textbf{0.077}   & \textbf{0.099}   & \textbf{0.094}  & \textbf{0.095}
  \\  \bottomrule
\end{tabular}
\label{tab:compare_hand_cabh}
\vspace{-0.5cm}
\end{table}

\vspace{-0.2cm}
\subsection{Evaluation on Multi-Dimensional Hand Trajectory Prediction}
\label{sec:eval_htp_md}

We first validate our claim that the proposed Uni-Hand predicts plausible hand center waypoints in both 2D and 3D dimensions. Predicting hand centers constitutes the most fundamental capability required by the HMF method.
The evaluation metrics including averaged displacement error (ADE) and final displacement error (FDE) are reported following the previous works \cite{bao2023uncertainty,ma2024diff,bao2024handsonvlm}. The prediction canvas~\cite{ma2024diff} is selected as the first frame of each video clip. That is, all the 3D waypoints are expressed relative to the first camera coordinate, and all the 2D waypoints are transformed to the first image plane.
The evaluation in the 3D space follows the absolute scale in meters. For the evaluation in 2D space, we normalize the predicted waypoint pixel coordinates by the image size. We separately forecast left- and right-hand motion. \myblue{We compare our Uni-Hand with the existing baselines including Constant Velocity Hand (CVH)~\cite{ma2024diff}, OCT~\cite{liu2022joint}, USST~\cite{bao2023uncertainty}, S-Mamba~\cite{wang2024mamba}, Diff-IP2D~\cite{ma2024diff}, EER (based on Diff-IP2D)~\cite{ma2025eer}, EMAG~\cite{hatano2024emag}, and MADiff~\cite{ma2024madiff}.} We modify S-Mamba~\cite{wang2024mamba} originally designed for general time series forecasting into our diffusion-based paradigm to predict HM tokens. \myblue{We additionally replace the 2D input and output waypoints, as well as the corresponding encoders and decoders with 3D counterparts in Diff-IP2D, EER, EMAG, and MADiff since they were originally developed for 2D forecasting tasks, obtaining the baselines Diff-IP3D, EER3D, EMAG3D, and MADiff3D.}

\begin{figure*}
  \centering
  \includegraphics[width=1\linewidth]{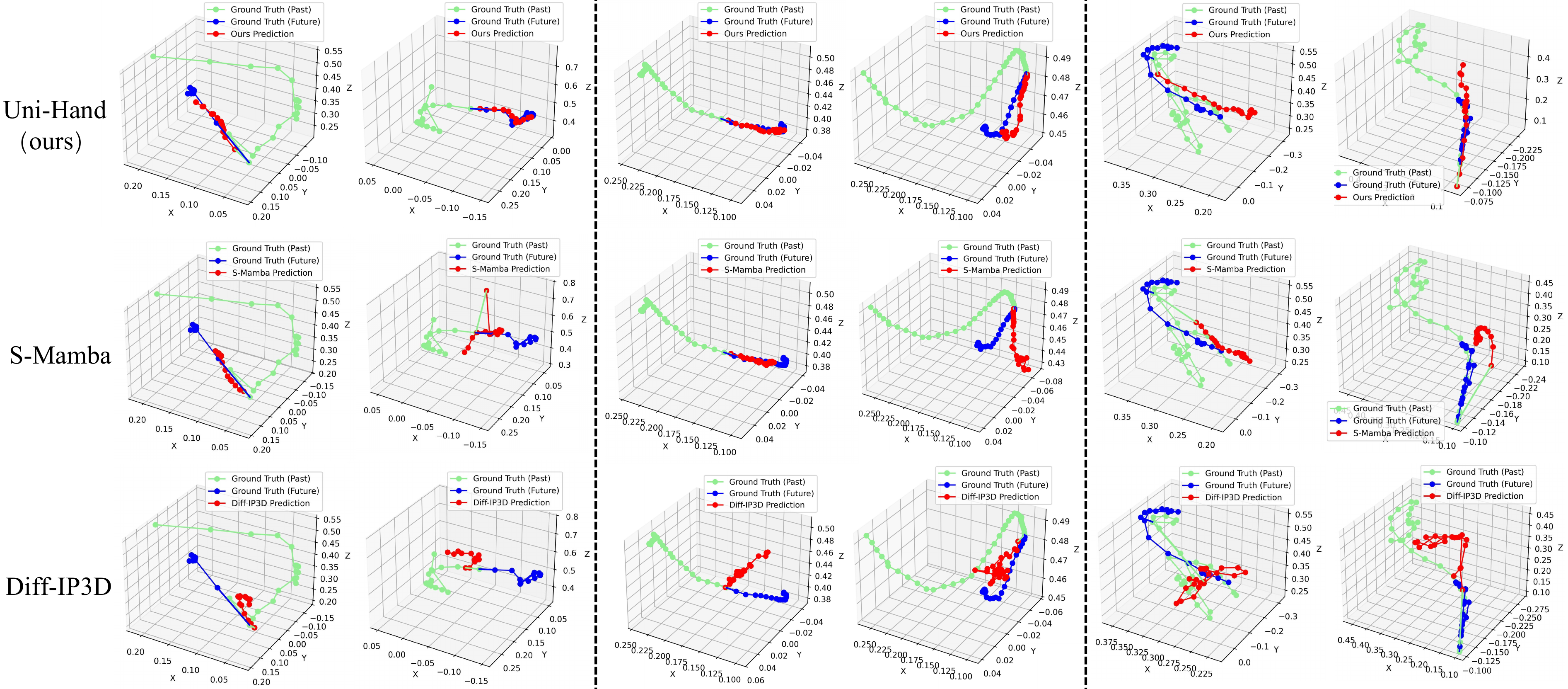}
  \caption{Visualization of predicted hand trajectories in the 3D space (left: EgoPAT3D-DT; middle: H2O-PT; right: HOT3D-Clips). We show the holistic sequence including observed past hand waypoints (green), ground-truth future ones (blue), and predicted future counterparts (red) by our Uni-Hand and two SOTA baselines.}
  \label{fig:viz_egopat_h2o_3d}
  \vspace{-0.2cm}
\end{figure*}

\begin{figure*}
  \centering
  \includegraphics[width=0.9\linewidth]{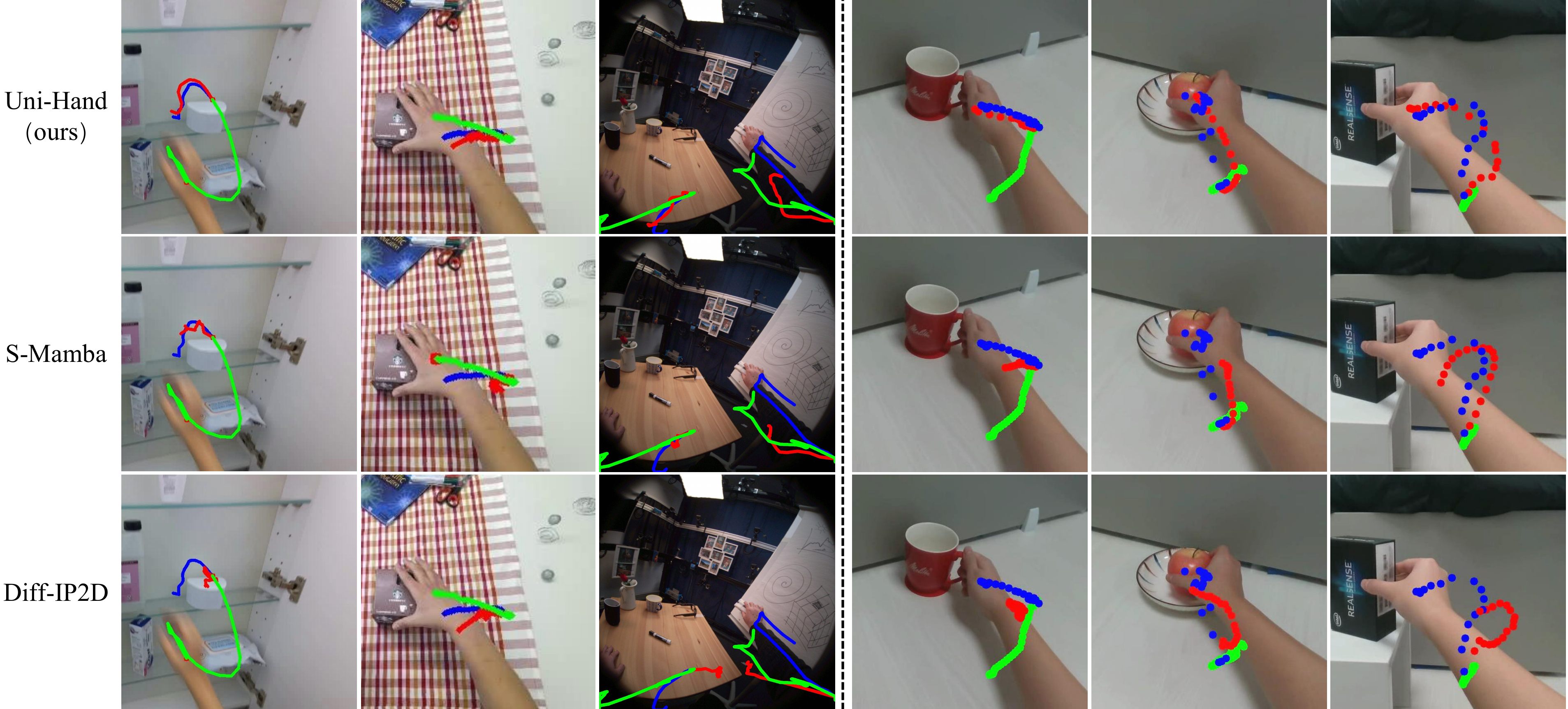}
  \caption{Visualization of 2D hand center forecasting (left: EgoPAT3D-DT, H2O-PT, and HOT3D-Clips; right: CABH). We show the holistic sequence including observed past joint waypoints (green), ground-truth future ones (blue), and predicted future counterparts (red) by our Uni-Hand and two SOTA baselines. Here we showcase the predictions on the first frame of each input video clip for EgoPAT3D-DT, H2O-PT, and HOT3D-Clips, while presenting the last frame for our CABH.}
  \label{fig:viz_2d_pred}
  \vspace{-0.3cm}
\end{figure*}

Here we present the center prediction performance of our proposed Uni-Hand and the baselines on the publicly available datasets and our CABH benchmark. As Tab.~\ref{tab:compare_hand_egopat_h2o_3d}$\sim$Tab.~\ref{tab:compare_hand_cabh} show, Uni-Hand produces the lowest prediction errors on most metrics compared to the SOTA baselines. Although some baselines show competitive performance in 2D and 3D respectively, they show specialized and lack consistent superiority across both spaces due to a lack of multi-modal fusion. In contrast, our universal framework accommodates both 3D structure information and 2D camera egomotion, achieving consistently comparable performance in both 2D and 3D spaces.
\myblue{Besides, Uni-Hand consistently outperforms the baseline methods on Ego4D, which indicates that Uni-Hand can more effectively forecast semantically driven hand motion under limited observation (only 3-frame input) and highly dynamic head motion, rather than only extrapolating short-term kinematic hand patterns.}
Fig.~\ref{fig:viz_egopat_h2o_3d} and Fig.~\ref{fig:viz_2d_pred} visualize the predicted hand center waypoints, where Uni-Hand's predictions have better stability and directionality in multiple dimensions. This suggests that Uni-Hand better captures hand motion patterns and potential human intentions during interaction by harmonizing multi-modal information. Moreover, HOT3D-Clips has small data volumes to supervise prediction models, where our Uni-Hand consistently outperforms the other baselines. This implies that our method better accommodates situations with limited training data.

\begin{table*}[t]
\small
\setlength{\tabcolsep}{20.2pt}
\center
\renewcommand\arraystretch{0.7}
\caption{Comparison of performance on multi-joint waypoint prediction in 3D space on the H2O-PT, HOT3D-Clips, and CABH-E. Best and secondary results are viewed in \textbf{bold black} and \mygreen{green} colors respectively.}
\begin{tabular}{l|cc|cc|cc}
\toprule
\multicolumn{1}{l|}{\multirow{2}{*}{Approach}}   & \multicolumn{2}{c|}{H2O-PT} & \multicolumn{2}{c|}{HOT3D-Clips}  & \multicolumn{2}{c}{CABH-E} \\ \cmidrule{2-7} 
\multicolumn{1}{c|}{}                                                                               & ADE\,$\downarrow$    & FDE\,$\downarrow$ & ADE\,$\downarrow$   & FDE\,$\downarrow$ & ADE\,$\downarrow$   & FDE\,$\downarrow$   \\ \cmidrule{1-7}                 
USST$^*$~\cite{bao2023uncertainty}  & 0.102  & 0.123	 & 0.162   & 0.180 	& 0.105	 & 0.122  \\ 
Diff-IP3D~\cite{ma2024diff} & 0.152   & 0.160    & 0.200   	& 0.204  & 0.093   & 0.110   \\ 
S-Mamba~\cite{wang2024mamba} & 0.046  & 0.087   & 0.200 	& 0.229	 & 0.099      &  0.130 \\ 
Uni-Hand w/o $X^\text{tar}$  & \mygreen{0.038}   & \mygreen{0.076}   & \mygreen{0.121}  	& \mygreen{0.151}	 & \mygreen{0.047}    & \mygreen{0.056} \\    
\rowcolor{lightgray}
Uni-Hand (ours)  & \textbf{0.036}   & \textbf{0.066}	  & \textbf{0.118}	  & \textbf{0.148}	& \textbf{0.038}   & \textbf{0.046} \\ \bottomrule
\end{tabular}
\label{tab:compare_hand_multi_3d}
\\ \vspace{-0.15cm}
\vspace{-0.3cm}
\end{table*}

\subsection{Evaluation on Multi-Target Hand Trajectory Prediction}
\label{sec:eval_htp_mt}

\begin{figure}
  \centering
  \includegraphics[width=1\linewidth]{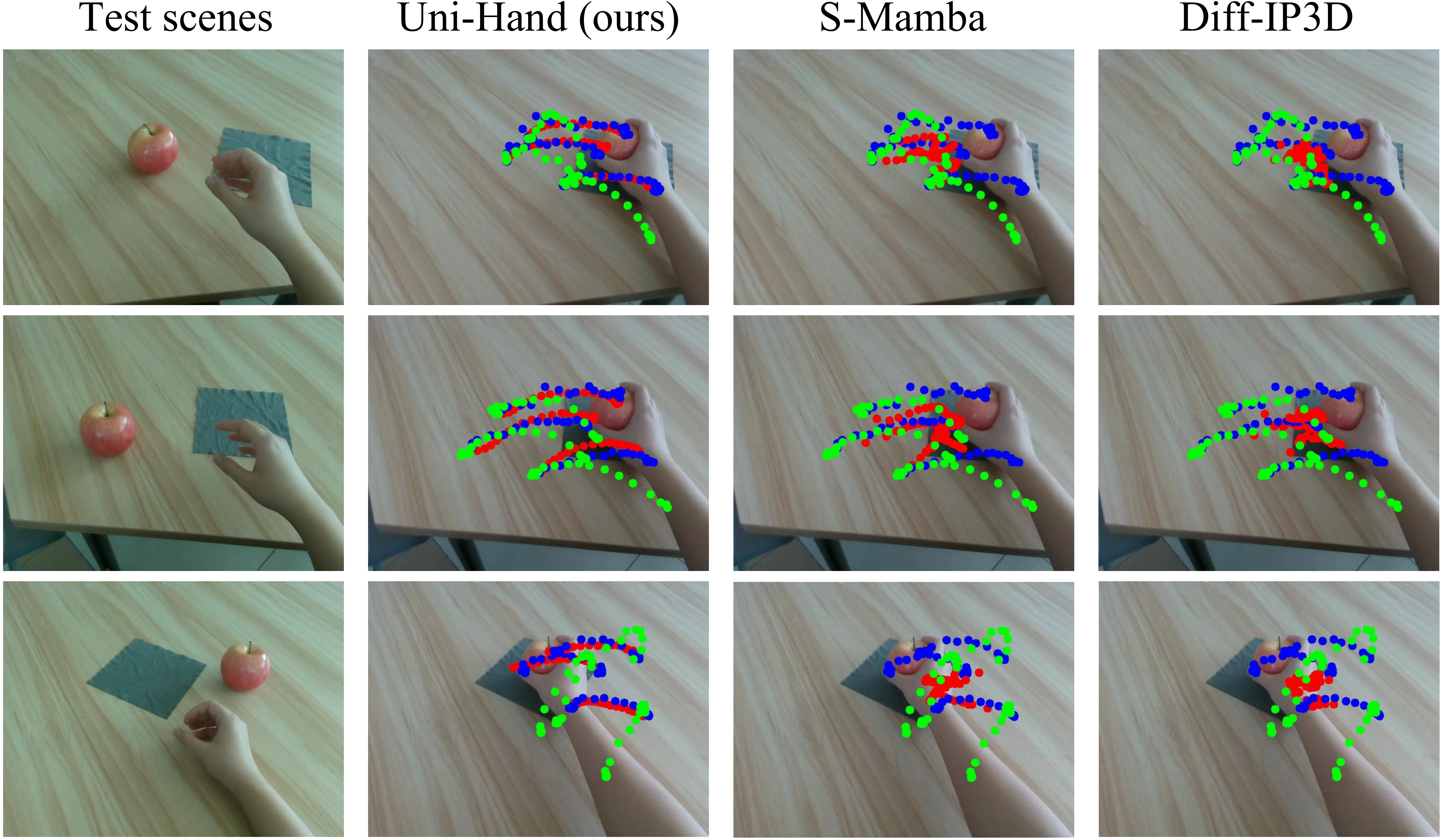}
  \caption{Visualization of multi-target prediction on our CABH-E. We show the holistic sequence including observed past joint waypoints (green), ground-truth future ones (blue), and predicted future counterparts (red) by our Uni-Hand and two SOTA baselines. Here we showcase the predictions on the last frame of each input video clip.}
  \label{fig:viz_cabh_e}
  \vspace{-0.55cm}
\end{figure}

We further validate Uni-Hand's capability to forecast plausible future waypoints of multiple hand joints rather than centers. We do not involve EgoPAT3D-DT here due to a lack of joint annotations. For H2O-PT and HOT3D-Clips which have labeled GT 3D joint positions, we predict trajectories of five joints: one on the wrist and two each on the thumb and index finger (see Supp. Mat., Sec.~B). These prediction targets are selected according to the recent work~\cite{ren2025motion} considering potential downstream tasks. Moreover, we provide a data extension of our CABH benchmark (CABH-E) compared to our preliminary version~\cite{ma2025mmtwin}. As shown in the first row of Fig.~\ref{fig:viz_cabh_e}, we record ``place the apple into the square area from multiple directions'' to demonstrate that Uni-Hand captures multimodal behavior~\cite{chi2023diffusion} without generating jittery actions from multiple directions. For CABH-E, we also generate 3D GT positions by HaMeR with depth observations, and forecast trajectories of three joints, which are on the wrist, the thumb tip, and the middle fingertip (see Fig.~\ref{fig:viz_cabh_e}). This target selection follows the setups from CoM~\cite{wang2025chain}, which emphasizes using the middle fingertips instead of the index ones. We thus showcase the performance on forecasting joint waypoints with different target combinations. The final displacement errors are averaged among the target joints. Note that the SOTA baselines in this experiment are implemented separately on different joints since they do not support multi-target prediction originally.

Tab.~\ref{tab:compare_hand_multi_3d} shows that Uni-hand consistently outperforms the baselines on the three datasets. Especially for CABH-E, Uni-hand reduces ADE and FDE of the second-best baseline by 0.055 and 0.064, respectively. We also visualize the predicted joints' positions projected to the last frame of each input video in Fig.~\ref{fig:viz_cabh_e}. Uni-Hand generally generates plausible joint waypoints extending in correct directions, with the final positions all successfully meeting the requirement of placing the apple in the target area. These results demonstrate that our universal framework achieves good multi-target hand forecasting, extending the scope of the existing hand trajectory prediction that only attends to hand centers.

Tab.~\ref{tab:compare_hand_multi_3d} also ablates the contribution of our devised target indicators. The baseline Uni-Hand w/o $X^\text{tar}$ separately forecasts different joints without using target indicators. The results indicate that although Uni-Hand w/o $X^\text{tar}$ has outperformed the other SOTA baselines, Uni-Hand can better identify which joint it currently attends to, and predicts more accurate joint waypoints. Therefore, the target indicators are demonstrated as key components in our universal HMF framework to achieve multi-target forecasting. \myblue{Please refer to Supp. Mat., Sec.~I for more ablations on target indicator formulations and conditioning mechanisms.}

\subsection{Evaluation on Downstream Tasks}
\label{sec:eval_downstream}

In this section, we evaluate the claim that Uni-Hand enables multi-task affordances for downstream applications.

\begin{figure*}
  \centering
  \includegraphics[width=0.92\linewidth]{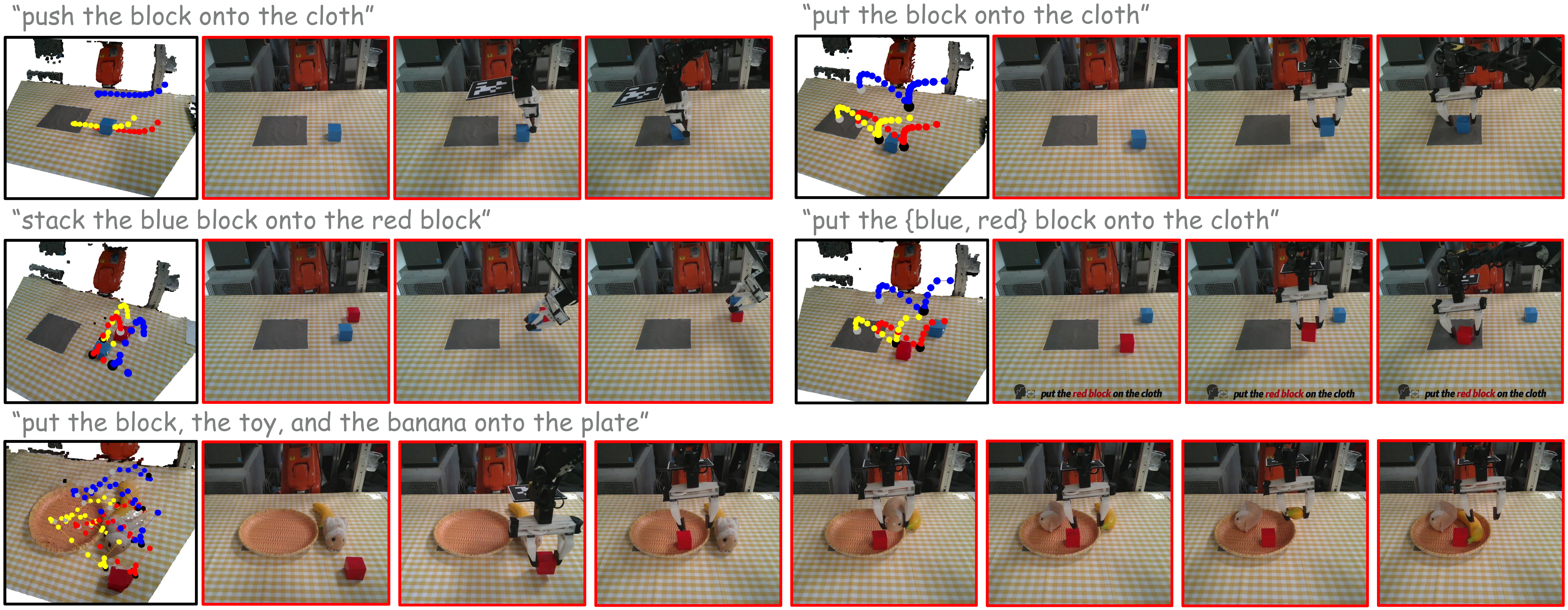}
    \vspace{-0.2cm}
  \caption{Real-world test on ALOHA. For each manipulation task, we illustrate the trajectories generated from scratch by Uni-Hand and how the gripper follows the action planning. The blue dots denote the predicted hand wrist positions, which spawn the yellow and red dots corresponding to the index fingertip and the thumb tip. The black and grey dots denote the pick and place timings of the gripper, respectively.}
  \label{fig:viz_real_aloha}
  \vspace{-0.2cm}
\end{figure*}

\subsubsection{Human-Robot Manipulation Policy Transfer}
\label{sec:eval_htp_mtpt}

This experiment is conducted on our proposed Hand-ALOHA-Transfer benchmark to validate Uni-Hand's ability to facilitate multiple downstream robotic manipulation tasks. The five tasks and prediction configurations in this benchmark have been detailed in Sec.~\ref{sec:hat_bench}. We activate the task-aware Transformer and inject the text embeddings of task instructions into it to enhance the model's adherence to specific task setups. Here, the GT interaction states are labeled manually by default. More experimental details and results with EgoLoc annotations will be introduced in Supp. Mat., Sec.~K. Tab.~\ref{tab:compare_hat} first shows 3D ADE and FDE metrics of Uni-Hand and the SOTA baseline S-Mamba on the validation sets of the real-world scenes. Uni-Hand generates end-effector trajectories that align better with human motion patterns. We also observed lower prediction errors around hand-object contact timings (see Supp. Mat., Sec.~P), which suggests that hand movements hold lower uncertainties at these timings, leading to more stable supervision signals to optimize Uni-Hand. This also facilitates plausible action planning despite relatively higher errors at other timings (e.g., the initial stage). Besides, we also present the Mean Absolute Error (MAE) of the contact-separation/separation-contact transition timings within the predicted interaction states on these tasks. Since the baseline~\cite{wang2024mamba} is proposed without the ability for interaction state prediction, we do not report its MAE and directly use GT labels as its pseudo predictions for the following real-world execution. Tab.~\ref{tab:compare_hat} shows that Uni-Hand predicts accurate hand-object contact/separation states for reasonable action planning. Then, we directly transfer the prediction models to the real-world deployment, and let the robot end-effector follow the generated trajectories and grasp actions. 

As shown in Tab.~\ref{tab:compare_hat}, for 10 test trials of each designated manipulation task, Uni-Hand exhibits significant superiority with higher success rates than S-Mamba. The baseline is infeasible with 0\% SR for the harder pick-and-place, language-conditioned pick-and-place, and long-horizon pick-and-place tasks. Fig.~\ref{fig:viz_real_aloha} shows how the end-effector accurately follows Uni-Hand's hand motion forecasting results for these tasks. This demonstrates that Uni-Hand provides the required motion knowledge for practical deployments on a real robot platform. It benefits policy learning for atomic skills such as push, pick, and place, and also holds promise for completing complex language-conditioned and long-horizon robotic tasks.

\begin{table*}[t]
\small
\setlength{\tabcolsep}{16.5pt}
\center
\renewcommand\arraystretch{0.7}
\caption{Performance comparison on our proposed HAT benchmark. We present the errors of hand trajectory prediction and interaction state prediction on the validation set (20 out of the training samples), as well as the success rates (SR) of 10 test trials for each manipulation task.}
\vspace{-0.2cm}
\begin{tabular}{c|cccc|cccc}
\toprule
\multicolumn{1}{c|}{\multirow{2}{*}{Tasks}}   & \multicolumn{4}{c|}{S-Mamba~\cite{wang2024mamba}} & \multicolumn{4}{c}{Uni-Hand (ours)} \\ \cmidrule{2-9} 
\multicolumn{1}{c|}{}                                                                               & ADE\,$\downarrow$    & FDE\,$\downarrow$ & MAE\,$\downarrow$  & SR\,$\uparrow$   & ADE\,$\downarrow$   & FDE\,$\downarrow$ & MAE\,$\downarrow$  & SR\,$\uparrow$    \\ \cmidrule{1-9}                 
\taskAA & 0.038  & 0.049 & -  & 40\%  & 0.022 & 0.018 & 1.3 & 100\% \\
\taskBB & 0.053 & 0.051 & - & 20\% & 0.035 & 0.012  & 1.9 & 90\%   \\ 
\taskCC & 0.061 & 0.040 & -  & 0\% & 0.040 & 0.016 & 0.8 & 80\%\\ 
\taskDD & 0.075  & 0.076 & - & 0\% & 0.054 & 0.028 & 1.8 & 80\%  \\
\taskEE & 0.100 & 0.109 & - & 0\% & 0.059 & 0.071 & 1.9 & 20\%  \\ \bottomrule
\end{tabular}
\label{tab:compare_hat}
\\ \vspace{-0.15cm}
\vspace{-0.2cm}
\end{table*}

\begin{figure*}
  \centering
  \includegraphics[width=1\linewidth]{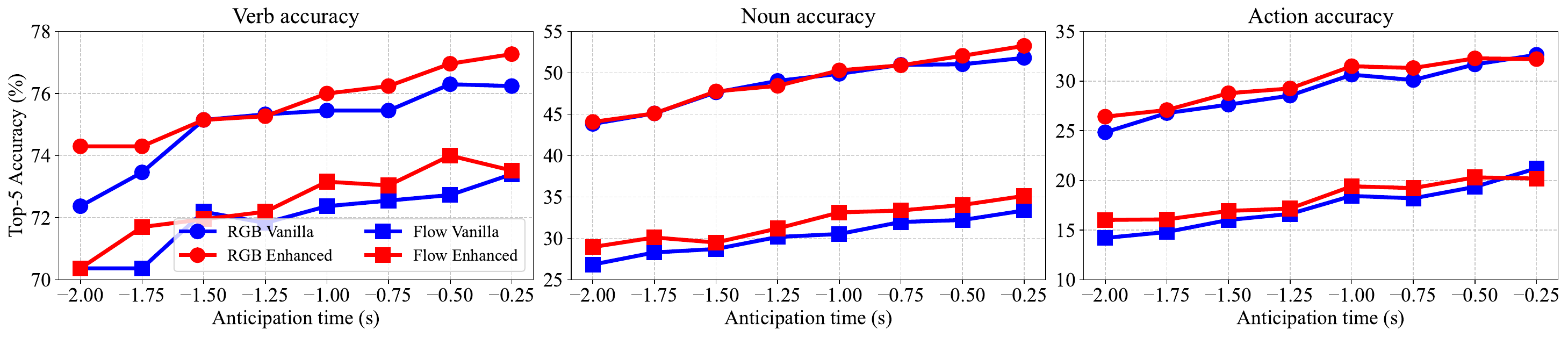}
  \vspace{-0.7cm}
  \caption{Performance comparison between RU-LSTM branches enhanced by our Uni-Hand's HM features and the vanilla model on the downstream action anticipation task. We follow the evaluation metrics of the prior work~\cite{furnari2020rolling}.}
  \label{fig:verb_noun_action_line}
  \vspace{-0.5cm}
\end{figure*}

\subsubsection{Action Anticipation}
\label{sec:eval_aa}
We have showcased in Fig.~\ref{fig:action_anticipation}(b) that our Uni-Hand can predict good hand waypoints after being optimized on Epic-Kitchens. This indicates that Uni-Hand can generate reasonable hand motion features (HM features) related to specific future human activities. Therefore, we explore whether the future HM features can enhance the downstream action anticipation, a task that predicts which category of action will happen.
As shown in Fig.~\ref{fig:verb_noun_action_line}, after we add the HM features recovered by our HMF diffusion to the vanilla modality-specific features in the RGB and flow branches of RU-LSTM, the Top-5 verb/noun/action accuracy increases at most anticipation times $\tau_a$. Fig.~\ref{fig:action_anticipation}(c) also presents an example of Top-5 action categories predicted by RU-LSTM before and after adding HM features. As can be seen, the feature enhancement by our HMF method facilitates a higher probability for the correct action category. These results highlight that the hand motion features predicted by Uni-Hand can be effective affordances encompassing future activity information for the downstream action anticipation.

\begin{table}[t]
\small
\setlength{\tabcolsep}{1pt}
\center
\renewcommand\arraystretch{0.7}
\caption{Top-5 action accuracy (\%) of RU-LSTM
branches on the downstream early action recognition and action recognition tasks. We present the performance at the observation rates 12.5\% $\sim$ 87.5\% for early action recognition, and 100\% for action recognition following~\cite{furnari2020rolling}. Best results are viewed in \textbf{bold black}.}
\vspace{-0.2cm}
\begin{tabular}{l|ccccccc}
\toprule
\multicolumn{1}{l|}{\multirow{2}{*}{Approach}}   & \multicolumn{7}{c}{Early action recognition} \\ \cmidrule{2-8} 
\multicolumn{1}{c|}{}    & 12.5\%    & 25.0\%  & 37.5\%   & 50.0\%  & 62.5\%    & 75.0\%  & 87.5\%  \\ \midrule                 
RGB (vanilla)  & \textbf{36.40}   & \textbf{38.21}   & 39.18   & \textbf{40.45} 	& 41.29	 & 41.90 & 41.66   \\ 
RGB (enhanced) & 36.22  & 37.85  & \textbf{39.42}  & 39.78  & \textbf{41.60}  & \textbf{42.26}  & \textbf{42.32}    \\  \cmidrule{1-8}   
Flow (vanilla)   &19.77  &\textbf{24.67}  &\textbf{27.63}  &28.30  &29.99  &30.35  &30.71         \\ 
Flow (enhanced) & \textbf{19.83}   & 24.43   & 26.90   & \textbf{28.54} 	& \textbf{30.29}  & \textbf{31.02} & \textbf{31.74} \\\midrule
\multicolumn{1}{l|}{\multirow{2}{*}{Approach}}   & \multicolumn{7}{c}{Action recognition (100\%)} \\ \cmidrule{2-8} 
\multicolumn{1}{c|}{} &\multicolumn{3}{c|}{Centers} & \multicolumn{4}{c}{Multiple joints}  \\ \cmidrule{1-8} 
RGB (vanilla) &\multicolumn{3}{c|}{41.78} & \multicolumn{4}{c}{41.78}  \\
RGB (enhanced) &\multicolumn{3}{c|}{\textbf{42.74}} & \multicolumn{4}{c}{\textbf{42.82}}  \\ \midrule
Flow (vanilla) &\multicolumn{3}{c|}{30.96} & \multicolumn{4}{c}{30.96}  \\
Flow (enhanced) &\multicolumn{3}{c|}{\textbf{31.74}} & \multicolumn{4}{c}{\textbf{31.86}}  \\ \bottomrule
               
\end{tabular}
\label{tab:compare_action_recognition}
\\ \vspace{-0.15cm}
\vspace{-0.1cm}
\end{table}

\subsubsection{Early Action Recognition and Action Recognition}
\label{sec:eval_ear}

Different from action anticipation, the downstream early action recognition and action recognition tasks focus on reasoning human activity categories according to partial or complete observations. Following RU-LSTM~\cite{furnari2020rolling}, we simply discard its ENCODING period and only attend to the observations in the ANTICIPATION period. As noted in Tab.~\ref{tab:compare_action_recognition}, after adding HM features from Uni-Hand, the recognition accuracy of RGB and flow branches in RU-LSTM increases at most observation rates.
Note that we additionally train a Uni-Hand model to achieve multi-joint motion forecasting (the wrist, the thumb tip, and the index fingertip), and enhance RU-LSTM by adding these new HM features. For action recognition with the observation rate of 100\%, HM features encompassing multi-joint motion information yield additional performance improvements. This further validates the advantages of multi-target predictions by Uni-Hand.
However, compared to the results of action anticipation in Fig.~\ref{fig:verb_noun_action_line}, the feature enhancement from Uni-Hand exhibits an overall weaker influence on these two downstream tasks. This suggests that the positive effect of predicted HM features diminishes when partial action observations are available. Nevertheless, Uni-Hand still affords informative HM features for action recognition thanks to a deeper understanding of hand motion patterns.

\myblue{Note that Uni-Hand is not intended as a standalone action anticipation/recognition model, but as an effective HMF framework whose predicted motion features can be directly plugged into existing anticipation/recognition models. Given that RU-LSTM has already been extensively optimized in the original work~\cite{furnari2020rolling}, consistent performance gains indicate that Uni-Hand provides meaningful future hand affordances, while larger improvements would require redesigning the downstream anticipation architecture.}

\begin{table}[t]
\small
\setlength{\tabcolsep}{4pt}
\center
\renewcommand\arraystretch{0.7}
\caption{Ablation study on camera egomotion. SE(3) as egomotion represents the baseline replacing the input camera homography with 6-DOF poses. Uni-Hand w/o EMF represents the baseline without the EMF diffusion. Best results are viewed in \textbf{bold black}.}
\vspace{-0.2cm}
\begin{tabular}{l|cc|cc|cc}
\toprule 
\multicolumn{1}{l|}{\multirow{2}{*}{Approach}}   & \multicolumn{2}{c|}{\makecell{EgoPAT3D-DT\\(seen)}} & \multicolumn{2}{c|}{\makecell{EgoPAT3D-DT\\(unseen)}}  & \multicolumn{2}{c}{H2O-PT} \\ \cmidrule{2-7} 
\multicolumn{1}{c|}{}                                                                               & ADE\,$\downarrow$    & FDE\,$\downarrow$ & ADE\,$\downarrow$   & FDE\,$\downarrow$ & ADE\,$\downarrow$   & FDE\,$\downarrow$   \\ \cmidrule{1-7}      
SE(3)  & 0.257  & 0.446   & 0.217    & 0.308   & 0.032   & 0.063     \\ 
w/o EMF  & 0.186 & 0.363  & 0.137  & 0.231  & 0.031  & 0.053   \\ 
\rowcolor{lightgray}
Uni-Hand  &\textbf{0.170}  &\textbf{0.336}	  &\textbf{0.118} 
  &\textbf{0.189} 	&\textbf{0.030} 	&\textbf{0.050}  \\ \bottomrule
\end{tabular}
\label{tab:abla_egomotion}
\\
\vspace{-0.5cm}
\end{table}

\subsection{Ablation Studies}
\label{sec:exp_albation}

\begin{figure}[t]
  \centering
  \includegraphics[width=1\linewidth]{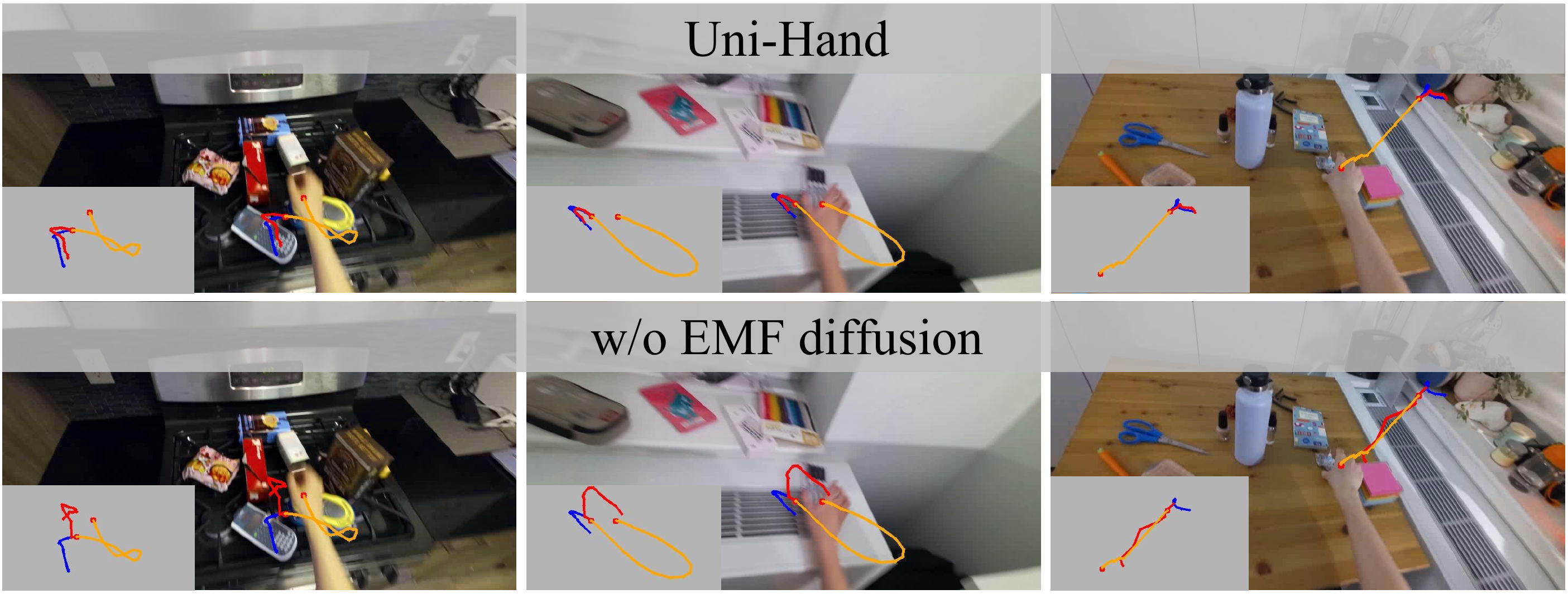}
  \caption{\myblue{Failure cases of the Uni-Hand's variant without future egomotion forecasting on EgoPAT3D-DT. The observed past hand waypoints, predicted future hand waypoints, and GT labels are represented in green, red, and blue, respectively, with the first frame of each sequence as the visualization canvas.}}
  \label{fig:viz_emf_ablate}
  \vspace{-0.3cm}
\end{figure}

\textbf{Camera Egomotion Forecasting.} We first ablate the headset camera egomotion forecasting on EgoPAT3D-DT and H2O-PT. Concretely, we conduct a baseline by removing the EM Diffusion, and regarding the last past camera homography as the constant egomotion in future time horizons.  
We present the hand center forecasting performance in 3D space in Tab.~\ref{tab:abla_egomotion}. As can be noted, forecasting future headset camera egomotion improves hand prediction accuracy. This demonstrates that our proposed dual-branch diffusion model in Uni-Hand captures the synergy between head motion and hand motion for a better understanding of the future interaction process. 
\myblue{Besides, Fig.~\ref{fig:viz_emf_ablate} visualizes the failure cases of the variant without camera egomotion forecasting in 2D space. As can be seen, explicitly forecasting future camera egomotion provides strong spatiotemporal constraints on hand trajectory prediction, leading to improved directional consistency and temporal stability.}

\textbf{Camera Egomotion Representation.} Despite the ablation studies on homography as camera egomotion provided by the previous works~\cite{ma2024diff}, we additionally present 3D hand center forecasting performance when we regard SE(3), i.e., 6-DOF poses from visual odometry, as camera egomotion instead of homography. 
As shown in Tab.~\ref{tab:abla_egomotion}, replacing the vanilla egomotion homography with SE(3) leads to a significant degradation in HMF performance. This suggests that the homography-based representation is better suited for modeling egomotion in this context. It also aligns with the fact that observed hands are constrained to 2D image planes, and homography matrices more effectively capture camera egomotion variations coupled with hand motions.

\begin{table}[t]
\small
\setlength{\tabcolsep}{1.65pt}
\center
\caption{Ablation study on multimodal inputs. Best results are viewed in \textbf{bold black}.}
\vspace{-0.2cm}
\renewcommand\arraystretch{0.7}
\begin{tabular}{cccc|cc|cc}
\toprule
\multicolumn{4}{c|}{Input modalities}  &\multicolumn{2}{c|}{Seen} &\multicolumn{2}{c}{Unseen}  \\ \midrule
waypoint & image & text & point cloud  & ADE\,$\downarrow$ & FDE\,$\downarrow$ & ADE\,$\downarrow$  & FDE\,$\downarrow$   \\ \midrule
\ding{51}      &       &     &         &0.178	&0.356	&0.124	&0.205  \\
 \ding{51}             & \ding{51}    &    &        &0.173	&0.350	&0.122	&0.201  \\
\ding{51}            & \ding{51}    & \ding{51}    &   &0.171 	&0.347	 &0.122	  &0.200\\ 
\rowcolor{lightgray}
\ding{51}            & \ding{51}    &\ding{51}  &\ding{51}     &\textbf{0.170}	&\textbf{0.336}	&\textbf{0.118}	&\textbf{0.189}  \\ \bottomrule
\end{tabular}
\label{tab:ala_on_inputs}
\vspace{-0.3cm}
\end{table}

\textbf{Multi-Modal Inputs.} To ablate multiple modalities, we incrementally add past hand waypoints, RGB images, the universal text prompt, and point clouds in model inputs. The experimental results of 3D hand center forecasting on EgoPAT3D-DT shown in Tab.~\ref{tab:ala_on_inputs} indicate that our proposed framework well harmonizes multi-modal information to achieve strong forecasting performance. The incorporation of both the universal text prompt and point cloud data effectively bridges modality gaps in existing literature.

\textbf{Text Embedding Injection.} 
We have shown in Fig.~\ref{fig:text_diff} that our text embedding injection in hybrid Mamba-Transformer facilitates the HM features denoised for task-aware HMF. Here we further present the quantitative results on the validation and test sets of the task ``put the \{\textit{blue}, \textit{red}\} block onto the cloth'' of our HAT benchmark. We conduct the baseline, Uni-Hand w/o TAT, by removing the task-aware Transformer in the denoising model. Tab.~\ref{tab:ala_on_tei} indicates that our text embedding injection with TAT reduces prediction errors, while also improving success rates significantly.

\begin{table}[t]
\small
\setlength{\tabcolsep}{7.5pt}
\center
\caption{Ablation study on text embedding injection. Best results are viewed in \textbf{bold black}.}
\vspace{-0.2cm}
\begin{tabular}{l|cccc}
\toprule
Approach   & ADE\,$\downarrow$    & FDE\,$\downarrow$ & MAE\,$\downarrow$    & SR\,$\uparrow$    \\ \cmidrule{1-5}    
Uni-Hand w/o TAT    & 0.057    & 0.034  & 2.0   & 40\%            \\
\rowcolor{lightgray}
Uni-Hand   & \textbf{0.054} & \textbf{0.028} & \textbf{1.8} & \textbf{80\%}            \\ \bottomrule
\end{tabular}
\label{tab:ala_on_tei}
\vspace{-0.5cm}
\end{table}

\vspace{-0.3cm}
\myblue{\section{Discussion} \label{sec:discussion}}

\myblue{Uni-Hand introduces a new algorithmic paradigm for hand motion forecasting that extends the existing approaches in multiple aspects, including modeling structure, prediction targets, conditioning objectives, and output space. At the modeling level, Uni-Hand is the first framework to jointly forecast future head egomotion and hand motion via a coordinated dual-branch diffusion architecture, enabling hand motion prediction to be explicitly conditioned on future egomotion. Beyond purely kinematic forecasting, Uni-Hand expands the output space by jointly predicting hand trajectories and hand-object contact/separation states, reasoning not only about where the hand will move but also when interactions will occur. Moreover, Uni-Hand introduces devised target indicators that enable a single predictive model to forecast heterogeneous hand targets, including the hand center, wrist, and finger joints, within a unified denoising process. The framework further incorporates global 3D scene context through a voxel-grid-based formulation, leading to 3D structure awareness under real-world geometric constraints. Ultimately, Uni-Hand injects task-aware textual embeddings directly into the diffusion denoising process through the dedicated cross-attention mechanism, aligning future hand motion generation with high-level task intentions. As can be noted, these algorithmic contributions redefine hand motion forecasting as a multi-target, interaction-aware, and task-conditioned generative problem, which we hope will inspire future research on egocentric human-object interaction.}\\

\section{Conclusion}
\label{sec:conclusion}

We have developed a universal framework namely Uni-Hand to achieve multi-dimensional and multi-target hand motion forecasting in egocentric views. Uni-Hand absorbs multi-modal data and predicts plausible hand trajectories in both 2D and 3D spaces, considering hand-head motion synergy. It attends to the future motion of hand centers and any joints, as well as hand-object interaction states. Extensive experiments on public datasets and our newly proposed benchmarks have demonstrated that our universal framework outperforms the existing SOTA baselines in hand motion forecasting across 2D/3D spaces and multiple targets, and facilitates multi-task
affordances for downstream applications, including robotic manipulation, action anticipation, early action recognition, and action recognition. In the future work, we plan to introduce tactile sensing for contact-rich human-robot policy transfer~\cite{hao2026rl}.

\vspace{-0.3cm}

\bibliographystyle{unsrt}
\bibliography{main}

\vspace{-1.5cm}
\begin{IEEEbiography}[{\includegraphics[width=1in,height=1.25in,clip,keepaspectratio]{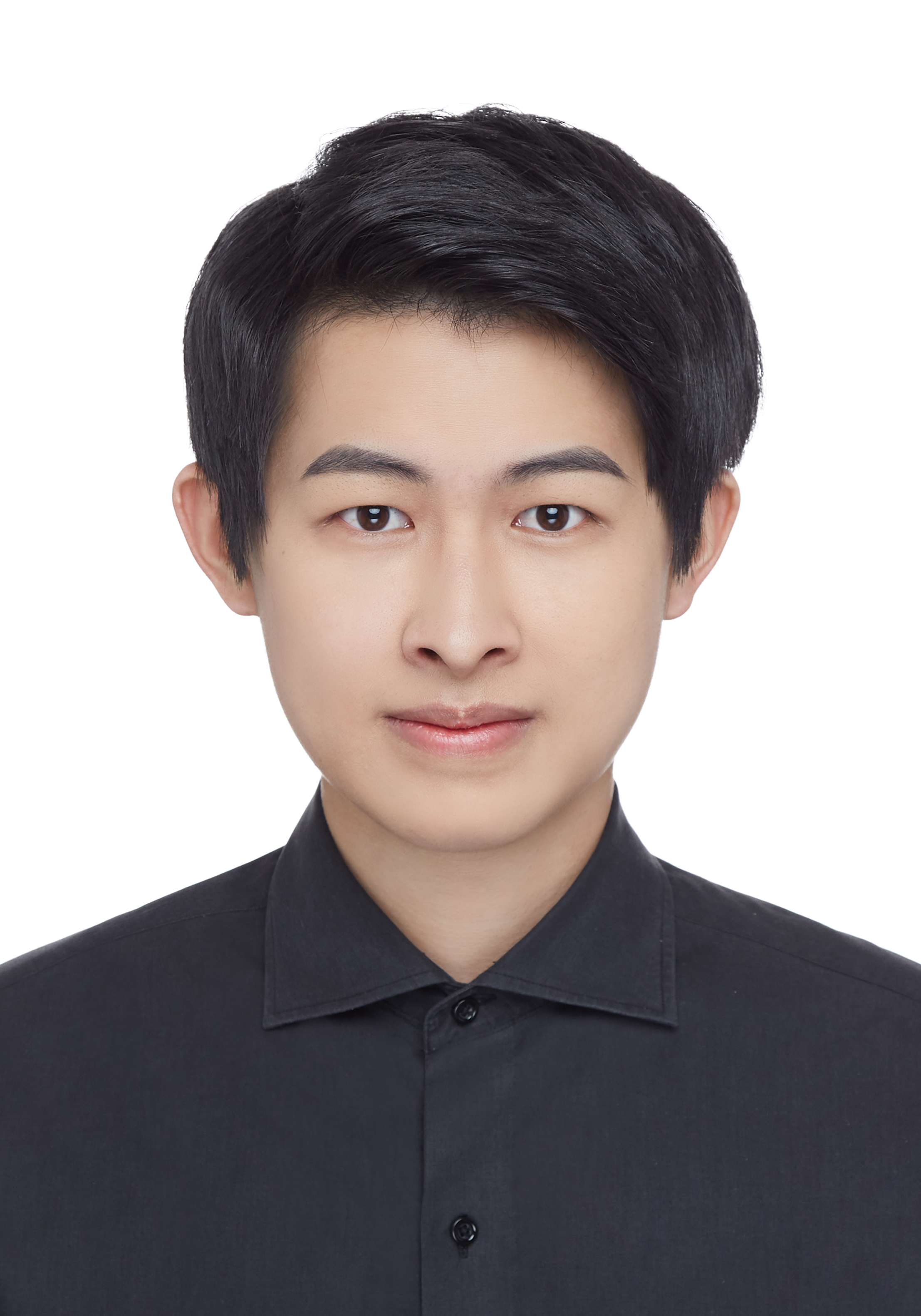}}]
{Junyi Ma} is a PhD candidate at IRMV Lab, Shanghai Jiao Tong University. He received his Master's degree and Bachelor's degree from the Beijing Institute of Technology, in 2020 and 2023 respectively. He is currently supervised by Prof. Hesheng Wang. His research interests include human motion analysis, human-robot skill transfer, and computer vision.
\end{IEEEbiography}
\vspace{-1.5cm}
\begin{IEEEbiography}[{\includegraphics[width=1in,height=1.25in,clip,keepaspectratio]{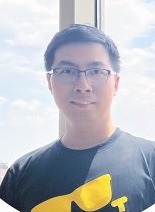}}]
{Wentao Bao} received Ph.D. degree from the Computer Science and Engineering Department at Michigan State University in USA, and BS and MS degrees from Wuhan University in China. His research focuses on open-world computer vision, including video activity recognition, prediction, and understanding in 2D/3D visual space. He has over 20 research works in leading CV/ML journals or conference venues such as CVPR, ICCV, ECCV, ACM MM, IEEE TIP, etc.
\end{IEEEbiography}
\vspace{-1.5cm}
\begin{IEEEbiography}[{\includegraphics[width=1in,height=1.25in,clip,keepaspectratio]{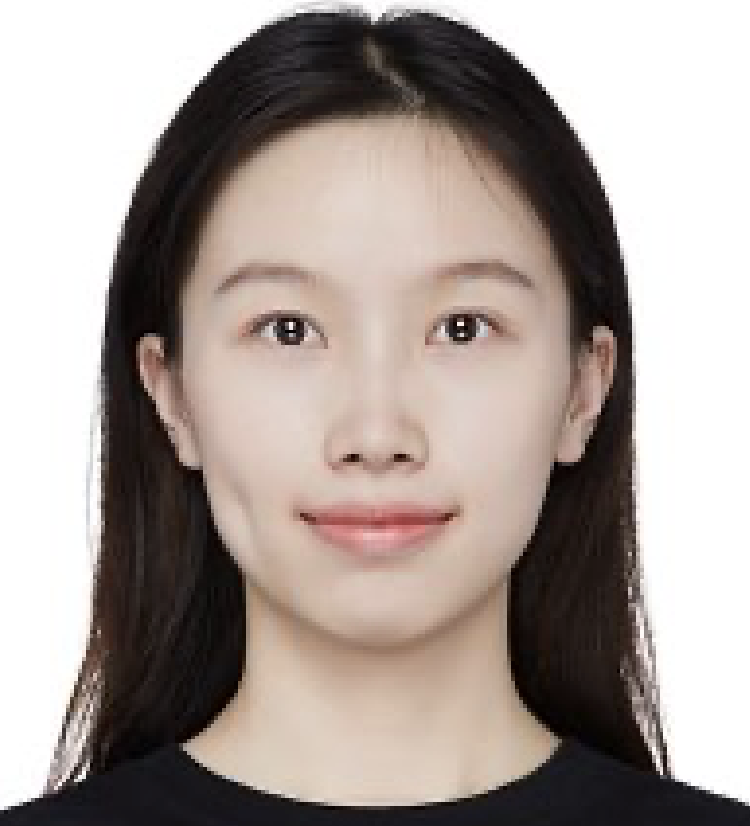}}]
{Jingyi Xu} is working toward the PhD degree from the Department of Electronic Engineering, Shanghai Jiao Tong University. She received her Master's degree and Bachelor's degree from the Beijing Institute of Technology, in 2020 and 2023 respectively. Her research interests are intelligent robots and autonomous driving. 
\end{IEEEbiography}
\vspace{-1.5cm}
\begin{IEEEbiography}[{\includegraphics[width=1in,height=1.25in,clip,keepaspectratio]{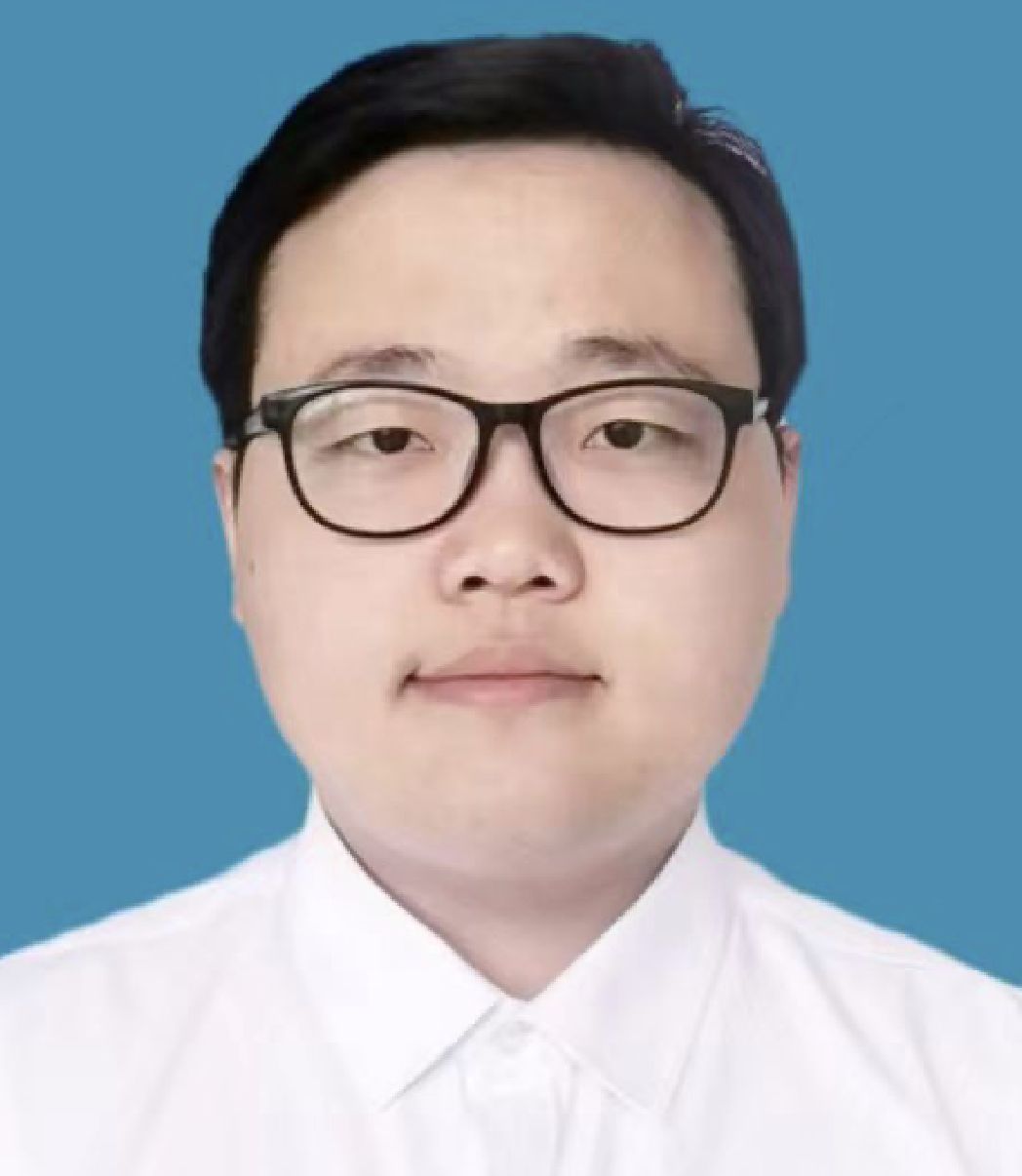}}]
{Guanzhong Sun} is a PhD candidate at IRMV Lab, China University of Mining and Technology. He received the Bachelor's degree from Nantong University in 2018 and received Master's degree from China University of Mining and Technology in 2022. He is currently supervised by Prof.Hesheng Wang. His research interests include vision-based manipulation and visual servoing. 
\end{IEEEbiography}
\vspace{-1.4cm}
\begin{IEEEbiography}[{\includegraphics[width=1in,height=1.25in,clip,keepaspectratio]{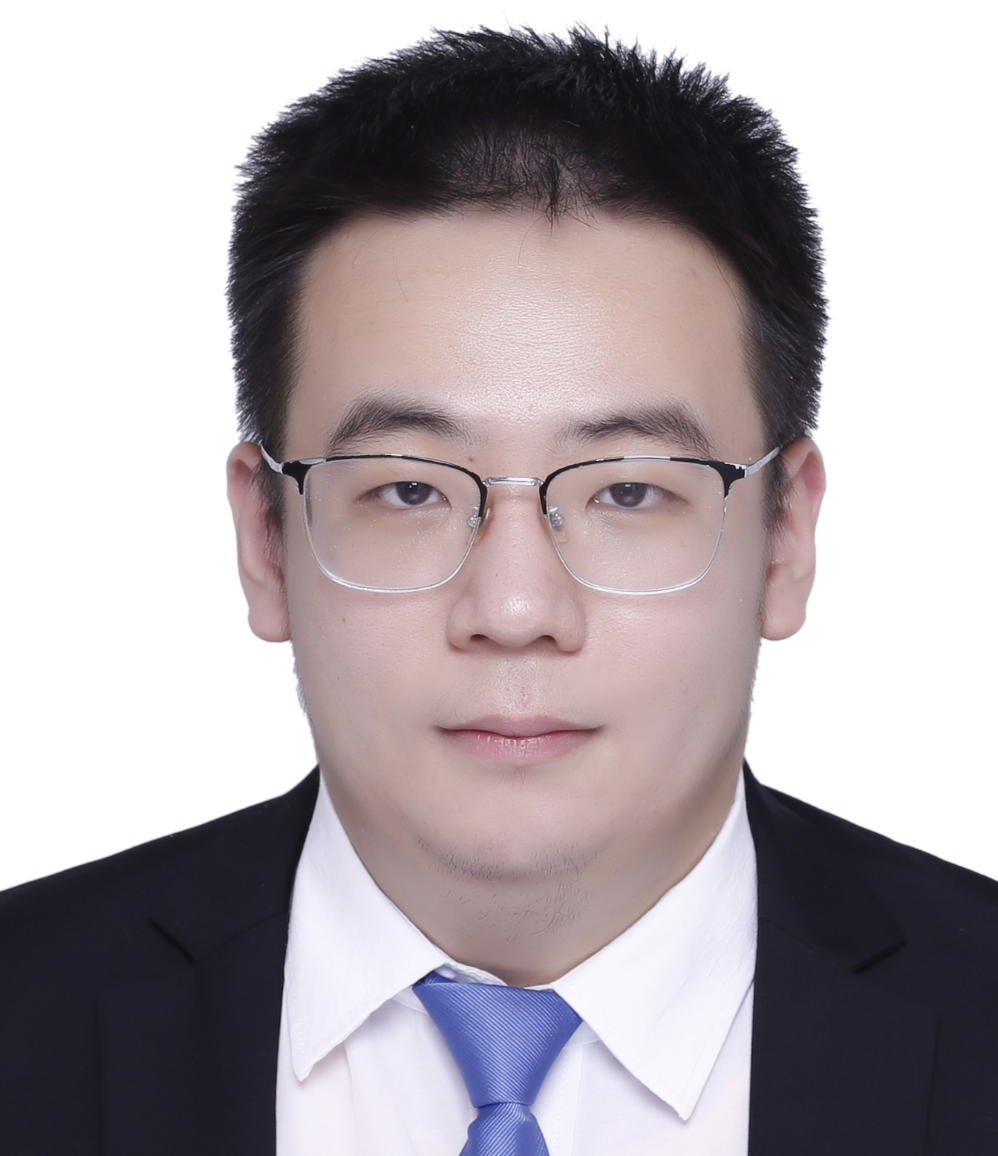}}]{Yu Zheng}
received the B.Eng degree in Department of Artificial Intelligence, Shanghai Jiao Tong University, China, in 2023. He is currently pursuing the Ph.D. degree in Control Science and Engineering with Shanghai Jiao Tong University, China. His current research interests include robot manipulation and computer vision.
\end{IEEEbiography}
\vspace{-1.5cm}
\begin{IEEEbiography}[{\includegraphics[width=1in,height=1.25in,clip,keepaspectratio]{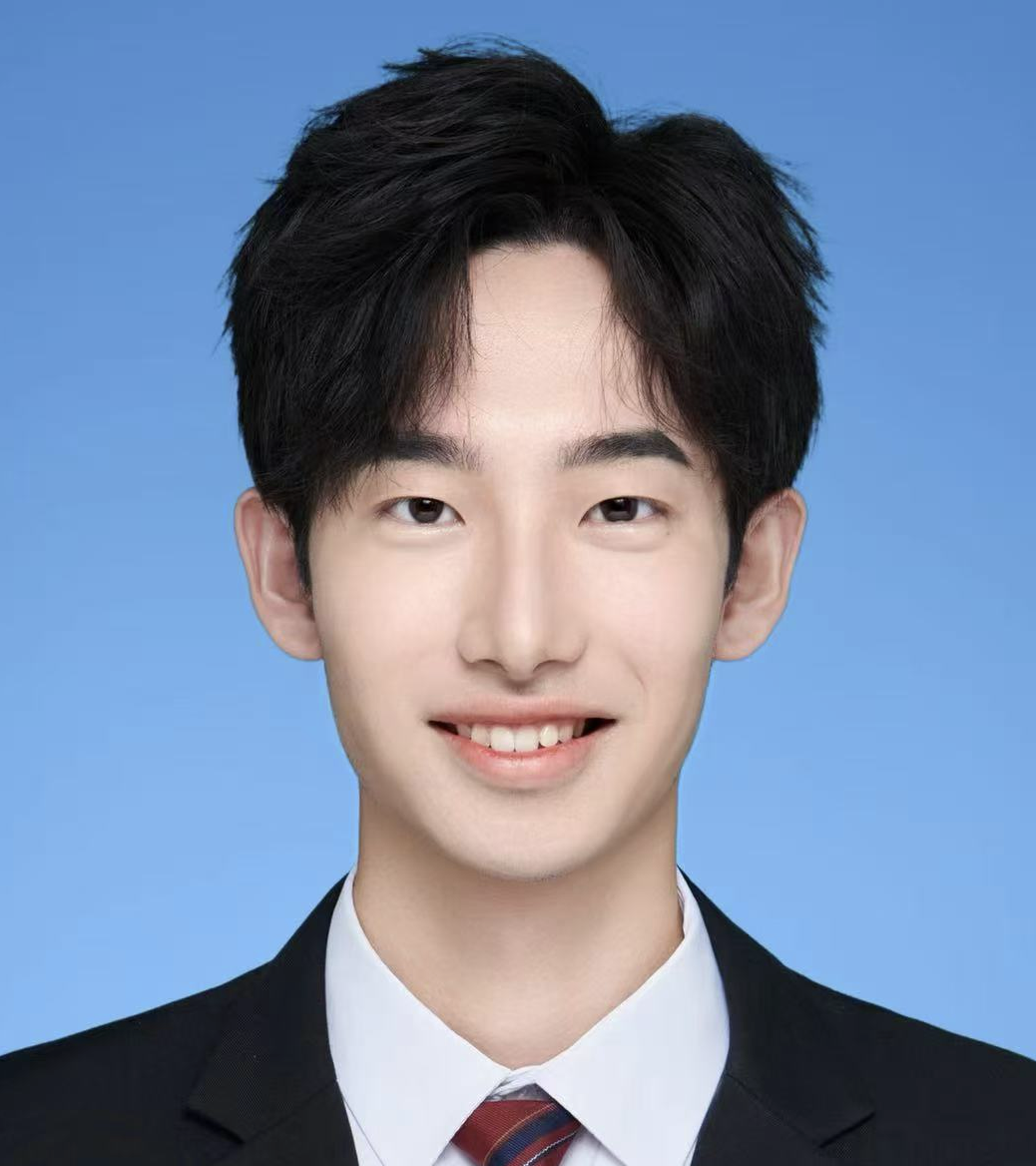}}]
{Erhang Zhang} is a PhD candidate at IRMV Lab, Shanghai Jiao Tong University. He received his Bachelor's degree from Shandong University in 2025. He is currently supervised by Prof. Hesheng Wang. His research interests include large-scale foundation models and embodied intelligence. 
\end{IEEEbiography}
\vspace{-1.5cm}
\begin{IEEEbiography}[{\includegraphics[width=1in,height=1.25in,clip,keepaspectratio]{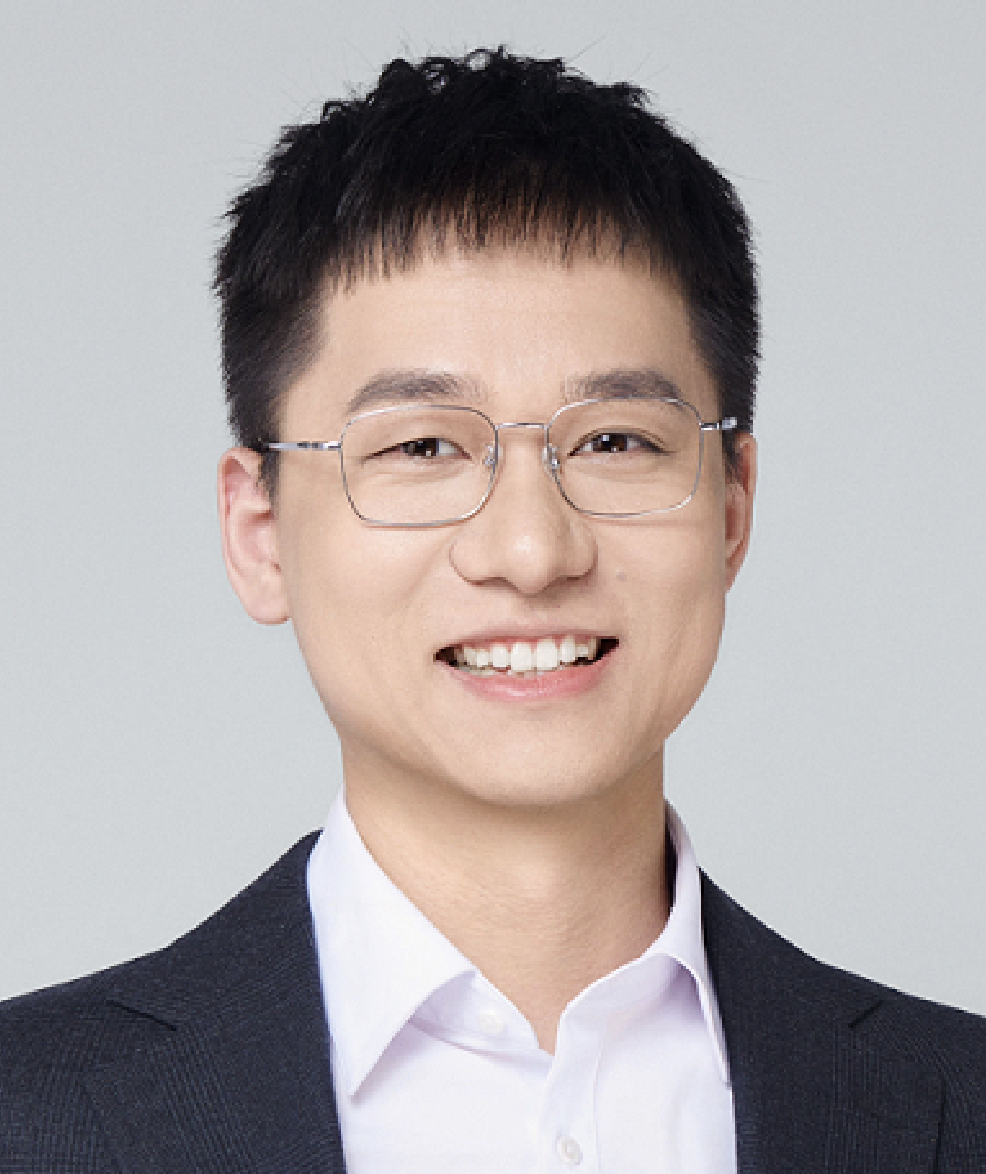}}]
{Xieyuanli Chen} is an Associate Professor at the National University of Defense Technology, China. He received his Ph.D. degree at the Photogrammetry and Robotics Laboratory at the University of Bonn. He received his Master's degree in Robotics in 2017 at the National University of Defense Technology. He received his Bachelor's degree in Electrical Engineering and Automation in 2015 at Hunan University. He serves as Associate Editor for IEEE RA-L, ICRA, and IROS. His research interests are SLAM, localization, mapping, and robot perception. 
\end{IEEEbiography}
\vspace{-1.1cm}
\begin{IEEEbiography}[{\includegraphics[width=1in,height=1.25in,clip,keepaspectratio]{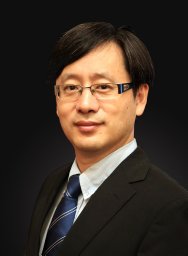}}]
{Hesheng Wang} (SM'15) received the B.Eng. degree in electrical engineering from the Harbin Institute of Technology, Harbin, China, in 2002, and the M.Phil. and Ph.D. degrees in automation and computer-aided engineering from the Chinese University of Hong Kong, Hong Kong, in 2004 and 2007, respectively. He is currently a Distinguished Professor with the School of Automation and Intelligent Sensing, Shanghai Jiao Tong University, Shanghai, China. His current research interests include visual servoing, intelligent robotics, computer vision, and autonomous driving. \\
Dr. Wang is an Associate Editor of Robotic Intelligence and Automation and the International Journal of Humanoid Robotics, a Senior Editor of the IEEE/ASME Transactions on Mechatronics, an Editor-in-Chief of Robot Learning. He served as an Associate Editor of the IEEE Transactions on Robotics from 2015 to 2019, an IEEE Transactions on Automation Science and Engineering from 2021 to 2023, He was the General Chair of IEEE/RSJ IROS 2025, IEEE ROBIO 2022 and IEEE RCAR 2016. 
\end{IEEEbiography}
\end{document}